\documentclass[sigconf]{acmart}
\usepackage{breqn}
\usepackage{graphicx}
\usepackage{mathrsfs}
\usepackage{multirow}
\usepackage{amsmath} 
\usepackage{setspace}
\usepackage{booktabs} 
\usepackage{caption}
\usepackage{subfigure}
\usepackage{balance}

\begin{document}
	
\title {Perceive Your Users in Depth: Learning Universal User Representations from Multiple E-commerce Tasks}

\author{Yabo Ni$^*$, Dan Ou$^*$, Shichen Liu, Xiang Li, Wenwu Ou, Anxiang Zeng, Luo Si}

\thanks{*Both authors contributed equally to this study}

\affiliation{%
  \institution{Search Algorithm Team, Alibaba Group, Seattle \& Hangzhou, China}
}
\email{{yabo.nyb, oudan.od, shichen.lsc, leo.lx, santong.oww, renzhong, luo.si}@alibaba-inc.com}

\copyrightyear{2018} 
\acmYear{2018} 
\setcopyright{acmcopyright}
\acmConference[KDD '18]{The 24th ACM SIGKDD International Conference on Knowledge Discovery \& Data Mining}{August 19--23, 2018}{London, United Kingdom}
\acmBooktitle{KDD '18: The 24th ACM SIGKDD International Conference on Knowledge Discovery \& Data Mining, August 19--23, 2018, London, United Kingdom}
\acmPrice{15.00}
\acmDOI{10.1145/3219819.3219828}
\acmISBN{978-1-4503-5552-0/18/08}	
\begin{abstract}

Tasks such as search and recommendation have become increasingly important for E-commerce to deal with the information overload problem. To meet the diverse needs of different users, personalization plays an important role. In many large portals such as Taobao and Amazon, there are a bunch of different types of search and recommendation tasks operating simultaneously for personalization. However, most of current techniques address each task separately. This is suboptimal as no information about users shared across different tasks.

In this work, we propose to learn universal user representations across multiple tasks for more effective personalization. In particular, user behavior sequences (e.g., click, bookmark or purchase of products) are modeled by LSTM and attention mechanism by integrating all the corresponding content, behavior and temporal information. User representations are shared and learned in an end-to-end setting across multiple tasks. Benefiting from better information utilization of multiple tasks, the user representations are more effective to reflect their interests and are more general to be transferred to new tasks. We refer this work as Deep User Perception Network (DUPN) and conduct an extensive set of offline and online experiments. Across all tested five different tasks, our DUPN consistently achieves better results by giving more effective user representations. Moreover, we deploy DUPN in large scale operational tasks in Taobao. Detailed implementations, e.g., incremental model updating, are also provided to 
address the practical issues for the real world applications.

\end{abstract}



\maketitle
\pagestyle{empty}  
\thispagestyle{empty} 
\section{Introduction}

In the Internet era, large portals such as Taobao and Amazon often contain hundreds of millions of items. It is difficult for users to find their desired items. Tasks such as search and recommendation have been utilized to address the information overload problem. Personalization techniques are critical for these tasks because better fitting personal needs can improve user experience and generate more business value.

There has been a large body of research studying personalization for recommendation~\cite{zhang2015daily}, search~\cite{wang2016learning, ustinovskiy2015optimization} and advertising~\cite{gopinath2010personalized}. Most of the research are based on matrix factorization (MF)~\cite{koren2009matrix} or neighborhood methods~\cite{linden2003amazon}. More recently, deep neural networks (DNNs) and recurrent neural networks (RNNs) are introduced for better performance. A lot of the previous works view the users as an item set (or sequence) with some side information, and aim to improve accuracy metrics such as Mean Absolute Error (MAE). Many of these techniques have been successfully used in real-world applications~\cite{covington2016deep, borisyuk2017lijar}. However, most these traditional personalization techniques focus on each individual task and build separate user models, which is sub-optimal as valuable user information is not shared across different tasks.

Different from most existing research, this paper focuses on the perception of portal users in a holistic manner, i.e., a representation which depicts a target customer in depth. The work aims at generating a general and universal user representation from multiple tasks on the portal, which can produce better personalization results for these tasks and can be transferred and utilized in new tasks. The new representation can extract general and effective features from complex user behavior sequences, and is shared and learned across several related tasks. In particular, this paper proposes the Deep User Perception Network (DUPN) that integrates the techniques of RNNs, attention and multi-task learning. User representations are modeled by the behavior sequences under different queries for more real-time session-based inference. RNNs are used as the building block to learn desired representations from massive user behavior logs. A novel attention network is designed on top of the RNN and learns to assign attention weights to items within the sequence by integrating all the corresponding content, behavior and temporal information to generate different user representations under different queries. By sharing representations and learning within the multi-task setting, the user representations are made more general and reliable.

An extensive set of experiments have been conducted with 4 tasks on the Taobao portal to learn the universal user representations. Evaluations on these 4 tasks show that the universal representations by multi-task learning can generate more accurate results than the setting of single-task learning. Furthermore, 1 new task was introduced and the corresponding results demonstrate the user presentations can be effectively transferred to the new task. Furthermore, the new research was deployed in the online e-commerce retrieval system of Taobao, and the Online A/B testing results suggest it can better reflect users' preference and generate more business value. Besides, the deployed model also has substantial advantage in efficiency. It is not necessary to build complex models from scratch for individual personalized tasks separately, which takes time and computing resources. In contrast, training models using the learned representations can be much simpler. Further more, realtime online inferring process of multi-tasks can be much faster and takes less online CPU cost, because the results of several tasks can be inferred once from the same network, instead of several big networks.

The contribution of the paper can be summarized as follows:

\begin{itemize}

\item{It proposes DUPN, a representation learning method based on multi-task learning, which enables the network to generalize universal user representations. We demonstrate that the shared representations can improve the performance of learned tasks, and can be transferred and applied successfully to other related tasks;

}

\item{It designs new attention and RNN based deep architecture for modeling users and items in e-commerce tasks as sequences of behaviors. A novel attention mechanism with query context is introduced to integrate all the corresponding content, behavior and temporal information for learning better vector representations of users;

}

\item{It thoroughly studies DUPN in both offline and online setting. In particular, the research was deployed in the operational system of Taobao search to capture users' real-time interest and recommend personalized results. Online A/B testing shows that the search results returned by our model meet users' requirements better and generate more business value.

}

\end{itemize}

The rest of the paper is organized as follows: Section 2 provides a literature survey of most related previous work. Section 3 gives the overview of the personalized ranking system of Taobao. Section 4 describes the proposed network architecture. Section 5 introduces the experimental methodology and Section 6 presents the experimental results and analysis. Section 7 provides some extra guidelines of training, maintaining and updating the network in Taobao search system. Section 8 concludes and points out some future research directions.

	\section{Related Work}
	There are mainly two lines of research that are related to our work: personalized recommendation with DNNs and RNNs, and multi-task representation learning. As techniques of e-commerce search and recommendation are similar, we survey both of them.

\subsection{Recommendation with DNNs and RNNs}
	
	In recent years, there are growing number of research on studying the recommendation and search personalization with deep neural networks. One of the earliest related public research with neural networks uses Restricted Boltzmann Machines (RBM) for Collaborative Filtering ~\cite{Salakhutdinov2007}. Salakhutdinov et al. model the user-item interaction with RBM and show its high competitiveness. In last several years, some research formulate Collaborative Filtering as deep neural networks and autoencoders. Wang et al. ~\cite{wanghao2015} employ neural network to extract features from content information of items and then use the extracted feature to enhance the recommendation performance. Sedhain et. al ~\cite{Suvash2016} and Yao et al.~\cite{yaowu2016} both model the Collaborative Filtering with antoencoders, where two fully connected layers are learned as encoder and decoder separately. Each user (item) is denoted as a vector, which is a row in the user-item rating matrix. The vector can be represented as itself with antoencoders. Meanwhile, network architectures have been well studied in literatures. Google proposes a wide and deep network~\cite{cheng2016wide} for advantages of both deep network and large scale sparse linear model. Microsoft proposes deep structured semantic models (DSSM) ~\cite{huang2013} where a cosine loss layer is introduced to make the model applicable on large-scale search applications. There are also some works using DNNs on specific domain recommendation. Neural networks are used for recommending music in ~\cite{Oord2013}, news in ~\cite{oh2014personalized}. Youtube illustrates their video recommender system in ~\cite{covington2016deep} by embedding the users, videos and queries in the portal. LinkIn proposes the job redistribution model based on a modified wide and deep network ~\cite{borisyuk2017lijar}.
	
	The  sequential characteristic of user behaviors on some applications makes RNNs a better choice  for recommendation. A sequence-to-sequence RNN model is applied to infer user's future intents based on their previous click behavior sequence in ~\cite{hidasi2015session}. Tan et al. ~\cite{tan2016improved} present several extensions to basic RNNs to enhance the performance of recurrent models for session-based recommendation. In the context of search-based online advertising, Zhai et al.~\cite{zhai2016deepintent} investigate an attention-based RNN to map both queries and ads to real valued vectors, achieving better advertising efficiency.

\subsection{Multi-task Representation Learning}
	The success of machine learning algorithms generally depends on data representation, because different representations can entangle and hide more or less different properties of variation behind the data ~\cite{bengio2013representation}. Representations for a single task deep network can only capture information needed by the task and other information of the data is totally discarded. In contrast, multi-task learning allows statistical strength sharing and knowledge transfer, thus the representations can capture more underlying factors. The hypothesis have been confirmed by a number of research, demonstrating the effectiveness of representation learning algorithms in scenarios of multi-task learning and transfer learning ~\cite{bengio2011expressive, krizhevsky2012imagenet, collobert2011natural}.
	
	In some of the multi-task learning research, tasks are divided into main task and auxiliary tasks. These works only care about performance on one task, and the other tasks are helping to learn better network structure or feature representations. In ~\cite{seltzer2013multi}, the network is trained to perform both the primary classification task and one or more auxiliary tasks using shared representations. This work shows that multi-task learning can provide a significant decrease in classification error rate. Zhang et al.~\cite{zhang2014facial} optimize facial landmark detection together with tasks of head pose estimation and facial attribute inference, arguing that the performance of the main task can be much better. In the other research, tasks are treated equally. In scenarios of drug discovery, Ramsundar et al. ~\cite{ramsundar2015massively} demonstrate that multi-task learning accuracy increases continuously with the number of tasks. Multi-task DNN approach of ~\cite{liu2015representation} is probably the most closely related work to ours. It combines tasks of domain classification and information retrieval, and demonstrates significant gains over the former task. Ranjan et al.~\cite{ranjan2016hyperface} design a method called HyperFace, simultaneously learn tasks of face detection, landmarks localization, pose estimation and gender recognition using  convolutional neural networks (CNN). Each task behaves significantly better than many competitive algorithms. Recently Chen et al. ~\cite{chen2017adversarial} propose adversarial multi-criteria learning for Chinese word segmentation (CWS). Each criteria is assigned with a task, and adversarial networks are employed to distinguish criteria of different dataset. Experiments show that joint learning on multiple corpus obtains a significant improvement compared to learning separately.
		
	To the best of our knowledge, our work is the first piece of study that uses deep multi-task representation learning to generate general and transferable user representations for personalization in operational e-commerce portal. 
	
	\section{System Overview}
\begin{figure}[t]
	\centering
	\includegraphics[width=0.95\linewidth]{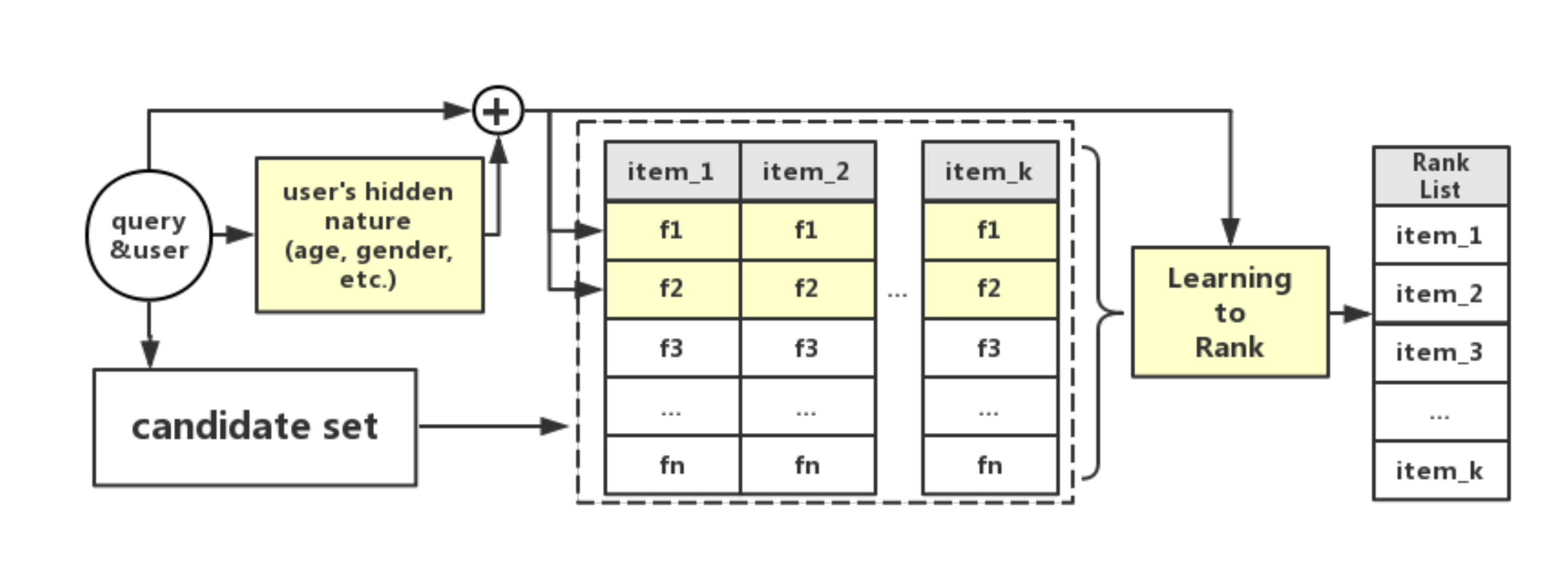}
	\caption{System Overview, $\{item_1,item_2,\ldots,item_k\}$ is the candidate item set recalled by current user and query, $\{f_1,f_2,\ldots,f_n\}$ is the ranking features for a corresponding item, $RankList$ is the final results presented to a user. Box in yellow denotes that it's a personalized task.}
	\label{fig:sys}
	\end{figure}

    The overall structure of our retrieval system is illustrated in Figure ~\ref{fig:sys}. The system maintains a collection of items. Given a query from the user, the system retrieves items whose title containing the query words, ranks the items, and presents the top ranked list to the user. Ranking personalization plays a key role in our retrieval system.

\par
	In e-commerce retrieval system, ranking  strategy is not only based on the relevance of the items with respect to the current query, user profile and item's quality also play an important role. Particularly, user profile and behaviors can be used in several personalized tasks hierarchically. First, we may use them to infer some hidden features of the user, such as price preference, favorite dressing style, etc. Then, the inferred hidden features, together with the original user profile, are used to produce some personalized ranking features, such as price-matching score, personalized CTR prediction, etc. Item' s quality is also modeled as features, e.g., item's history sales volume. Some of the ranking features in our retrieval system are listed in Table 1. There are tens of ranking features in our online retrieval system. A personalized learning to rank (L2R) model is finally trained to combine all these ranking features together. The L2R model is trained to learn the weights for different ranking features based on the current user and query, and returns the final rank list of items which maximizes the global conversion rate or other business targets.

\begin{table}[t]
	\centering
	\caption{Examples of Ranking Features}
	\newsavebox{\tablebox}
	\begin{lrbox}{\tablebox}
		\begin{tabular}{c|c}
			\hline
			Feature Name & Description \\\hline
			Sales Volume & The sales volume of a certain item \\\hline
			Rating Score &  Users' ratings on a certain item\\\hline
			\multirow{2}[-2]*{CTR}& A prediction of click through rate on items  \\
			Prediction& using a logistic regression model \\\hline
			{Price Preference} & How the item price matches the user\\
			\hline 	
		\end{tabular}
	\end{lrbox}
	\scalebox{0.9}{\usebox{\tablebox}}
	\label{Feature}
\end{table}

\par

	\section{model architecture}

\begin{figure}[t]
	\centering
	\includegraphics[width=1\linewidth]{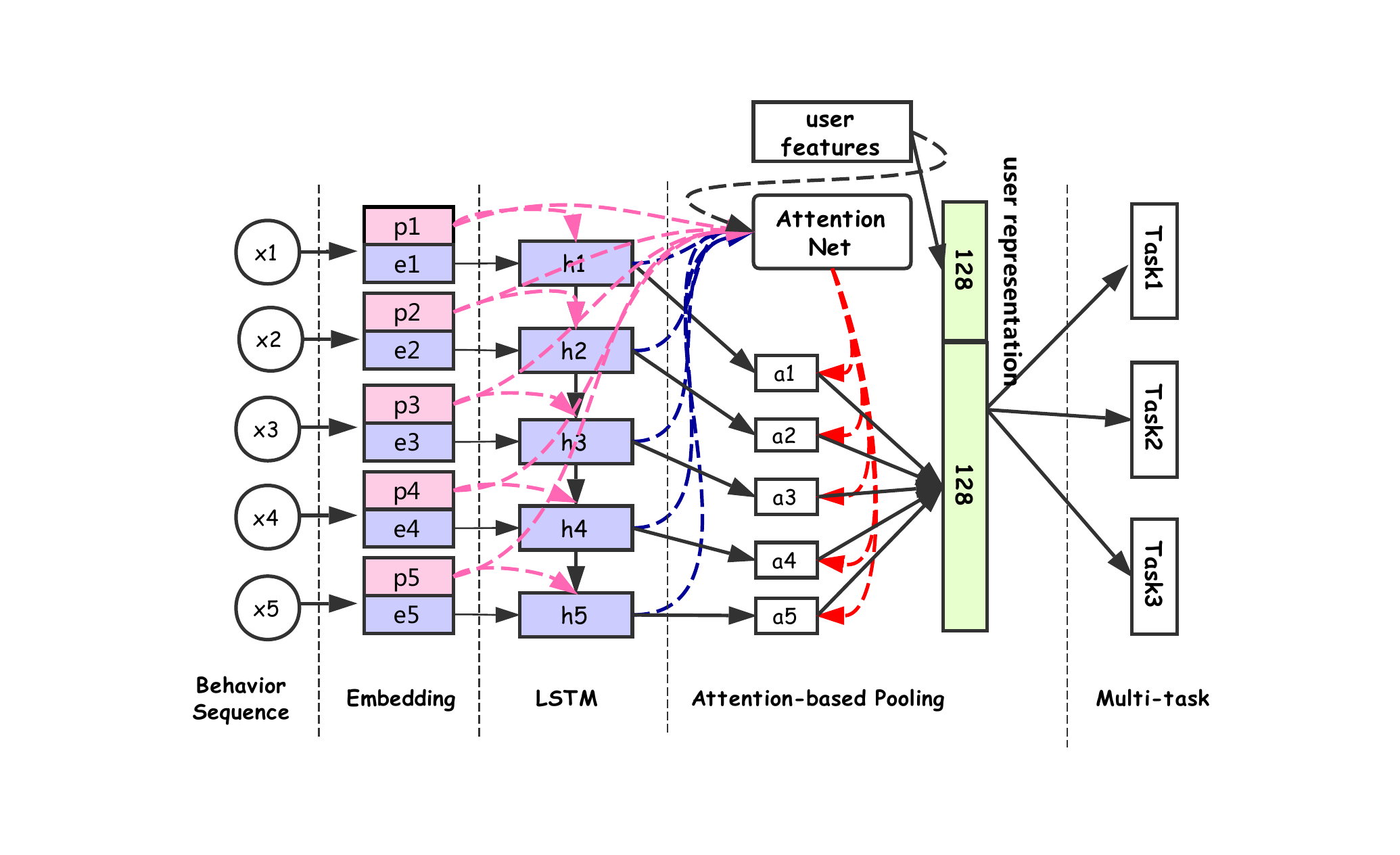}
	\caption{General Model Architecture. Color of lilac denotes the item, pink denotes the behavior properties and reseda denotes the user representation. This color style is consistent with all the figures of the network structure. }
	\label{fig:Model}
	\end{figure}

	Figure ~\ref{fig:Model} shows the general network architecture of DUPN. The model takes user behavior sequence as input and transfers each behavior into an embedded vector space. Then we apply LSTM~\cite{Hochreiter} and attention-based pooling to obtain a user representation vector. LSTM helps to model the user behavior sequence and attention net helps to draw information from the sequence by different weights. By sharing representations between related tasks, we can enable our model to generalize better both on our learning tasks and some new tasks. We will first introduce each component of the encoder, then discuss the settings of the multi-tasks.
	
	\subsection{The Input \& Behavior Embedding}

	The input of DUPN is a user behavior sequence $x=\{x_1,x_2,\ldots,x_N\}$ where $x_i$ indicates the $i^{th}$ behavior and is ordered by time. Each user behavior contains a user-item interaction, such as click, purchase, etc. So $x_i$ is described as a pair $\left \langle item_i, property_i \right \rangle$, where $item_i$ indicates the Taobao product corresponding to $i^{th}$ behavior, and $property_i$ describes the specific feature of the behavior.
\par
	Features of different scales are used to represent an item ($item_i$). Generalized features of shop ID, brand, category and item tags are used to model its common factors while personalized feature of item ID is used to model the unique factor. For long-tailed items and new items, generalized features will play the leading role. While for popular items, personalized features will dominate. Item tags include item attributes and statistical features.
	 
\par
	$property_i$ describes the specifics of a user behavior on a certain item, including behavior type, behavior scenario and behavior time. Behavior types include click, bookmark, add to cart and purchase. Behavior scenario denotes where the behavior happens, such as in search, recommender scenario or advertising. Behavior time indicates the time gap between the behavior and current search, workdays or weekends, in the morning or at night, etc.
	
\par
\begin{figure}[tb]
	\centering
	\includegraphics[width=1.0\linewidth]{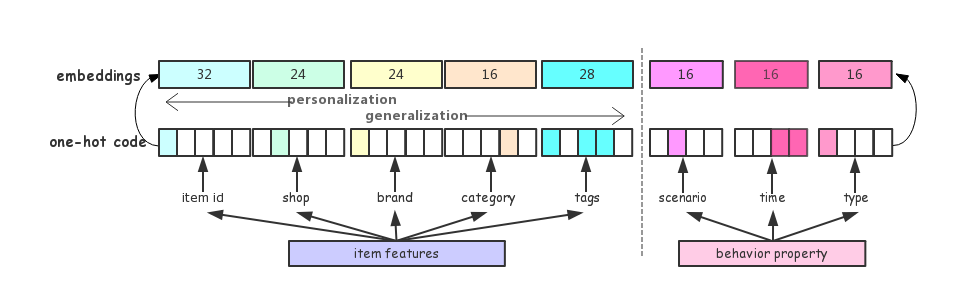}
	\caption{Behavior Embedding. Each behavior consists of an item and some properties of the behavior. }
	\label{fig:itemEmbedding}
	\end{figure}

	We model different features separately and then concatenate the embeddings to get the item representations. Each behavior $x_i$ is represented by a multi-hot vector $\{x_i^1,x_i^2,\ldots,x_i^F\}$ which represents item features (ID, category, brand, etc.) and the behavior property (scenario, type, time). $F$ is the number of features of each behavior, and $x_i^f$ is the $f^{th}$ feature which is a one-hot or multi-hot vector.
	The input layer and the embedding layer are illustrated in Figure ~\ref{fig:itemEmbedding}. For item $i$, the behavior embedding layer transforms the multi-hot vector $x_i=[item_i, property_i]$ into a low-dimensional dense vector $res_i = [e_i, p_i]$ ($e_i$ and $p_i$ represent the embedding of $item_i$ and $property_i$ respectively) by linear mapping, as is shown in Equation ~\ref{eq1}:

	\begin{equation}\label{eq1}
	res_i = [W_{emb}^1x_i^1,W_{emb}^2x_i^2,\ldots, W_{emb}^Fx_i^F], W_{emb}^f\in R^{d_{emb}^f\times V_f}
	\end{equation}
\par
	where $d_{emb}^f$ denotes the dimension of embedded vector of the $f^{th}$ feature and $V_f$ is the vocabulary size. The item embedding layer reduces the dimensionality of the item features from their vocabulary size to a much smaller size, thus regularizing the model. 
	Vocabulary size of items, brands, shops, categories and tags are 1G, 1M, 10M, 1M and 100K respectively. And the corresponding dimensions are 32, 24, 24, 16 and 28. The behavior property is embedded into 48 floats. To sum up, $res_i$ is a 172 dimensional vector.

\subsection{Property Gated LSTM \& Attention Net}
	
	Considering sequential characteristic of user behaviors in e-commerce, the output of the item embedding layer $\{res_1,res_2,\ldots,res_N\}$ is fed into LSTM. LSTM updates a hidden state $h_t\in R^{d_h}$ by the current input $res_t$ and the previous hidden state $h_{t-1} $in a recurrent formula. 
	Further considering traits of the two parts ($e_t$ and $p_t$) of a behavior ($res_t$), we propose a \textbf{Property Gated LSTM} where the behavior property $p_t$ and item feature $e_t$ are treated differently. Particularly, 1) $p_t$ cannot tell any speciality of the item or the user, but reflects the importance of each behavior, so we treat it as a very strong signal in the gates of LSTM (described in (2),(3) and(5)). In other words, what to extract, what to remember and what to forward are extensively affected by $p_t$. 2) $e_t$, which tells the item features and implies user interests, is the only input for LSTM  (described in (4)). The full Property Gated LSTM model is formulated as follows:
\begin{eqnarray}
	&i_t = &\sigma(W_{ei}e_t + W_{pi}p_t + W_{hi}h_{t-1} + b_i) \\
	&f_t = &\sigma(W_{ef}e_t + W_{pf}p_t + W_{hf}h_{t-1} + b_f) \\
	&c_t = &f_t\cdot c_{t-1} + i_t \cdot tanh(W_{ec}e_t + W_{hc}h_{t-1} + b_c)\\
	&o_t =&\sigma(W_{eo}e_t + W_{po}p_t + W_{ho}h_{t-1} + b_o)\\
	&h_t = &o_t \cdot tanh(c_t)
\end{eqnarray}
	
	\par
	where $i_t$, $f_t$ and $o_t$ represent the $input$, $forget$ and $output$ gates of the $t^{th}$ object respectively. $c_t$ is the cell activation vector. 
		
\begin{figure}[t]
	\centering
	\subfigure[Property Gated LSTM] {
  
		\begin{minipage}[b]{0.4\textwidth}
			\includegraphics[width=0.95\textwidth]{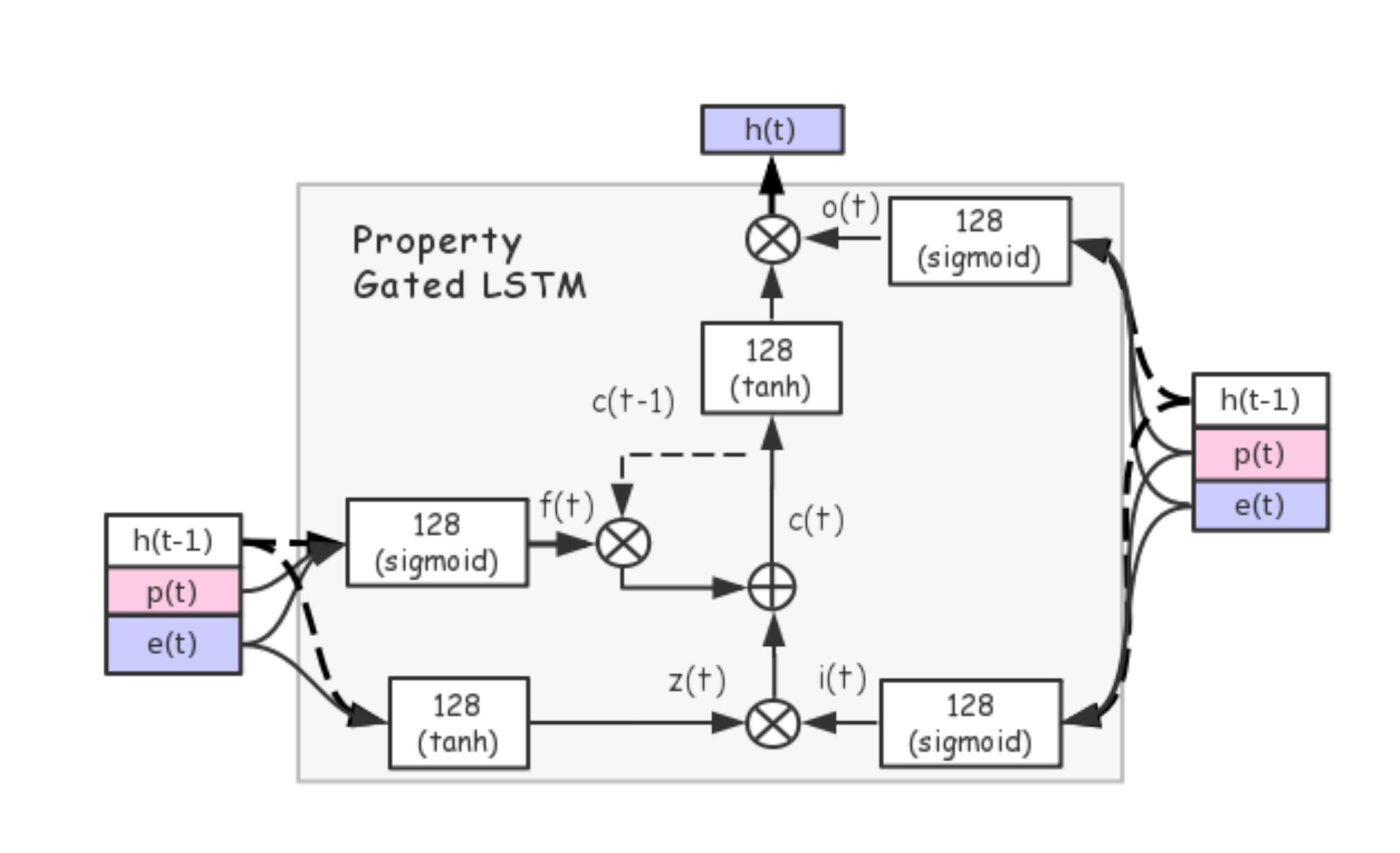}
		\end{minipage}
	}
	
	\subfigure[Attention Net] {
	
		\begin{minipage}[b]{0.4\textwidth}
			\includegraphics[width=0.95\textwidth]{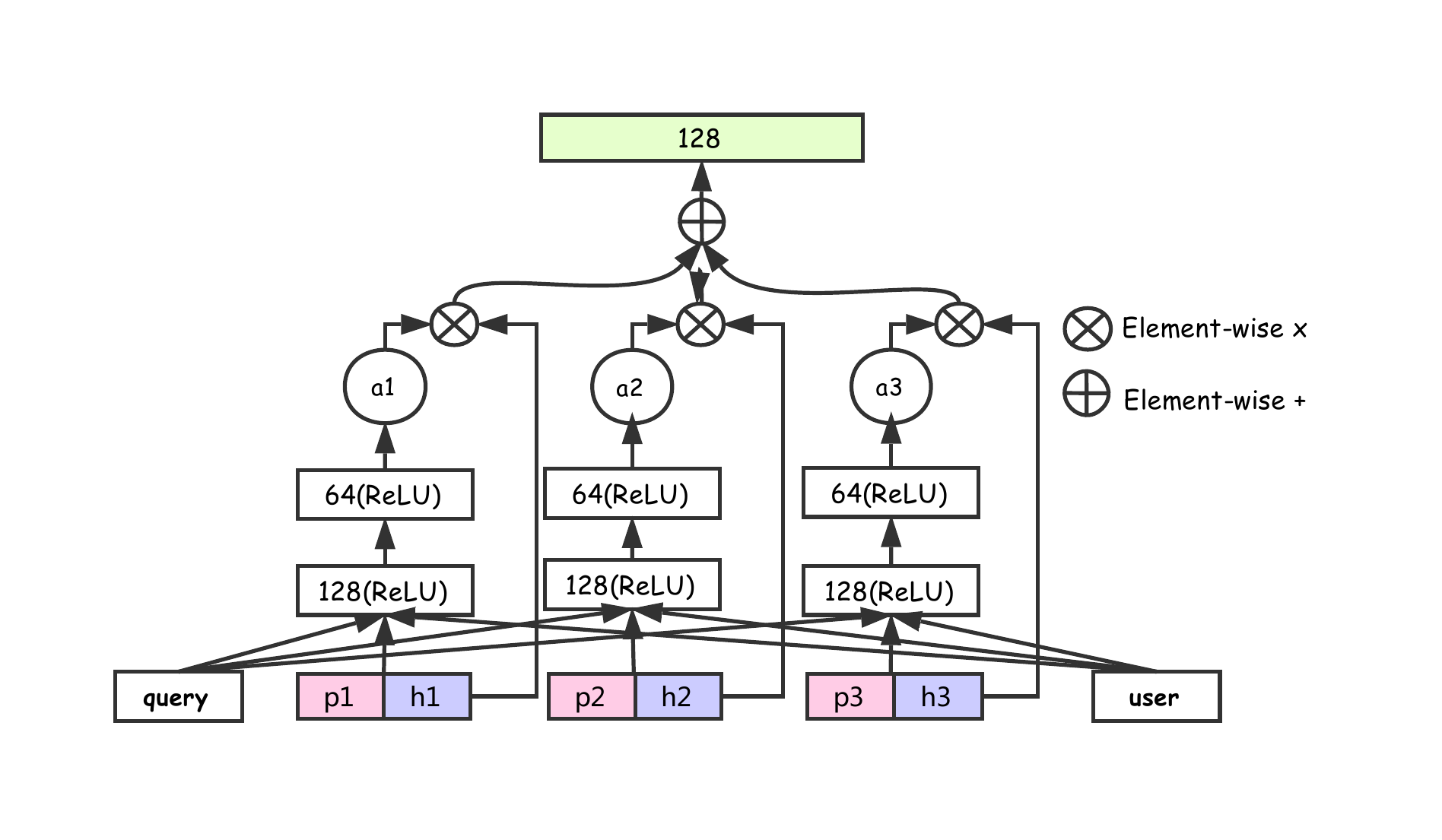}
		\end{minipage}
	}
	\caption{Property Gated LSTM and Attention Mechanism.}
	\label{fig:Attention Net}
	\end{figure}

	The output of Property Gated LSTM is another sequence $h=\{h_1,h_2,\ldots,h_N\}$. We apply an attention mechanism on the top of Property Gated LSTM. The attention net architecture is shown in Figure ~\ref{fig:Attention Net}(b). In the model, we consider each vector $h_i$ as the representation of $i^{th}$ item, and represent the sequence by a weighted sum of the vector representation of all the items. The attention weight makes it possible to perform proper credit assignment to items according to their importance to current query. Mathematically, it takes the form as follows:
	\begin{equation}
	\begin{split}
	&rep_s=\sum_{t=1}^Na_th_t ,\\
	&a_t=\frac{exp(attention(h_t,q,u,p_t;\omega))}{\sum_{t=1}^Texp(attention(h_t,q,u,p_t;\omega))}
	\end{split}
	\end{equation}
\par	
	where $a_t$ is the weight for each hidden state $h_t$, $attention(\cdot;\omega)$ is the attention net which is a two-layer fully connected net and takes query embedding $q$, user profile $u$, the $t^{th}$ hidden state $h_t$, and the behavior property $p_t$ as input. $rep_s$ is the representation of the sequence. The user representation $rep$ is the concatenation of $rep_s$ and $u$ (embedding from user profile), which is a 256 dimensional vector. Note that, similar to Property Gated LSTM, $p_t$ also contributes to the attention weight and $h_t$ is the objective of the attention mechanism.
	
	\subsection{Multi-tasks}
	\label{sec::behav}
	
\begin{figure*}[htbp]
	\centering
	\subfigure[CTR task, item embeddings are shared in DUPN]{
		\begin{minipage}[b]{0.19\textwidth}
			\includegraphics[width=0.985\textwidth]{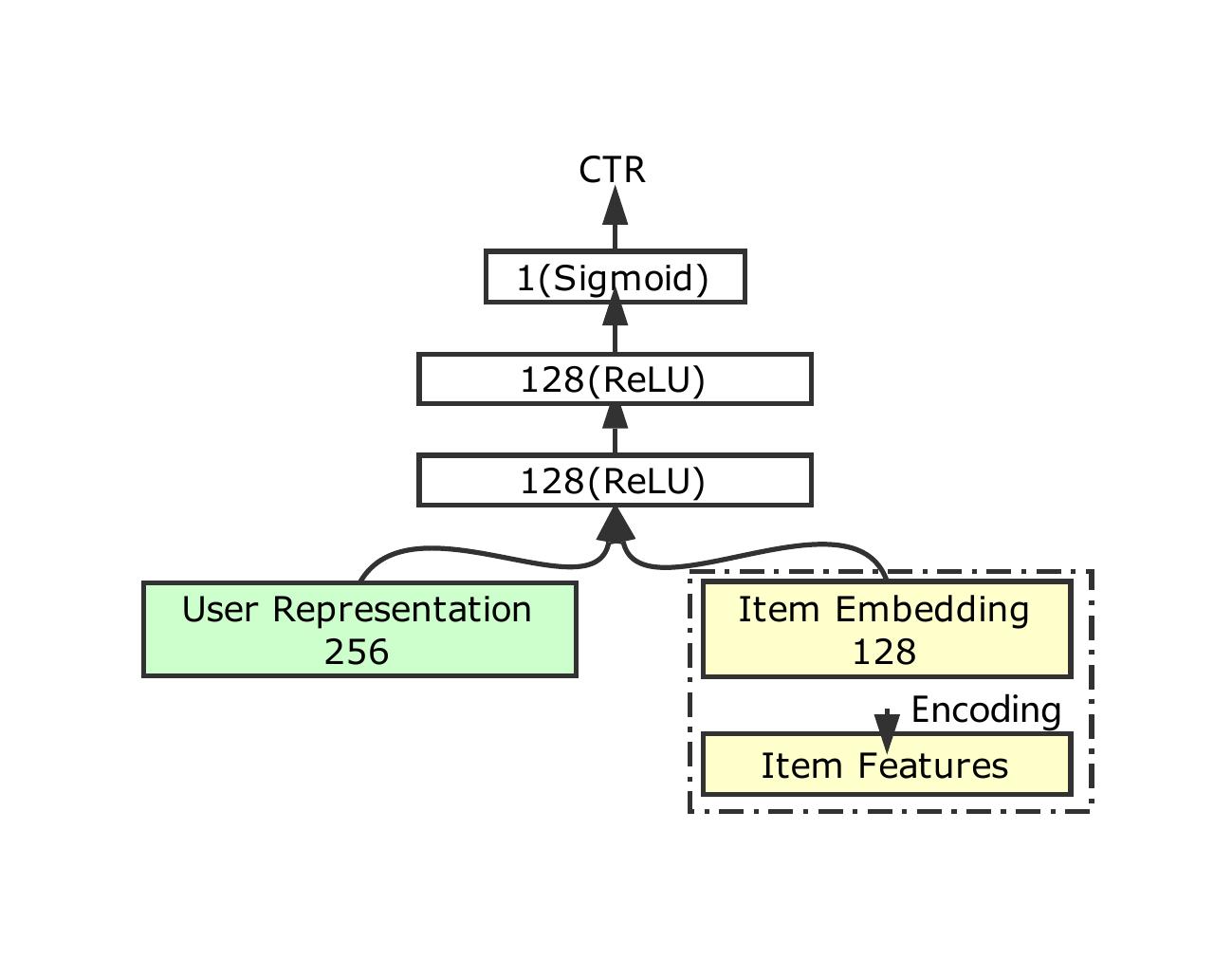} 
			\end{minipage}
	}
	\subfigure[L2R task, learning the ranking feature weights]{
		\begin{minipage}[b]{0.19\textwidth}
			\includegraphics[width=0.985\textwidth]{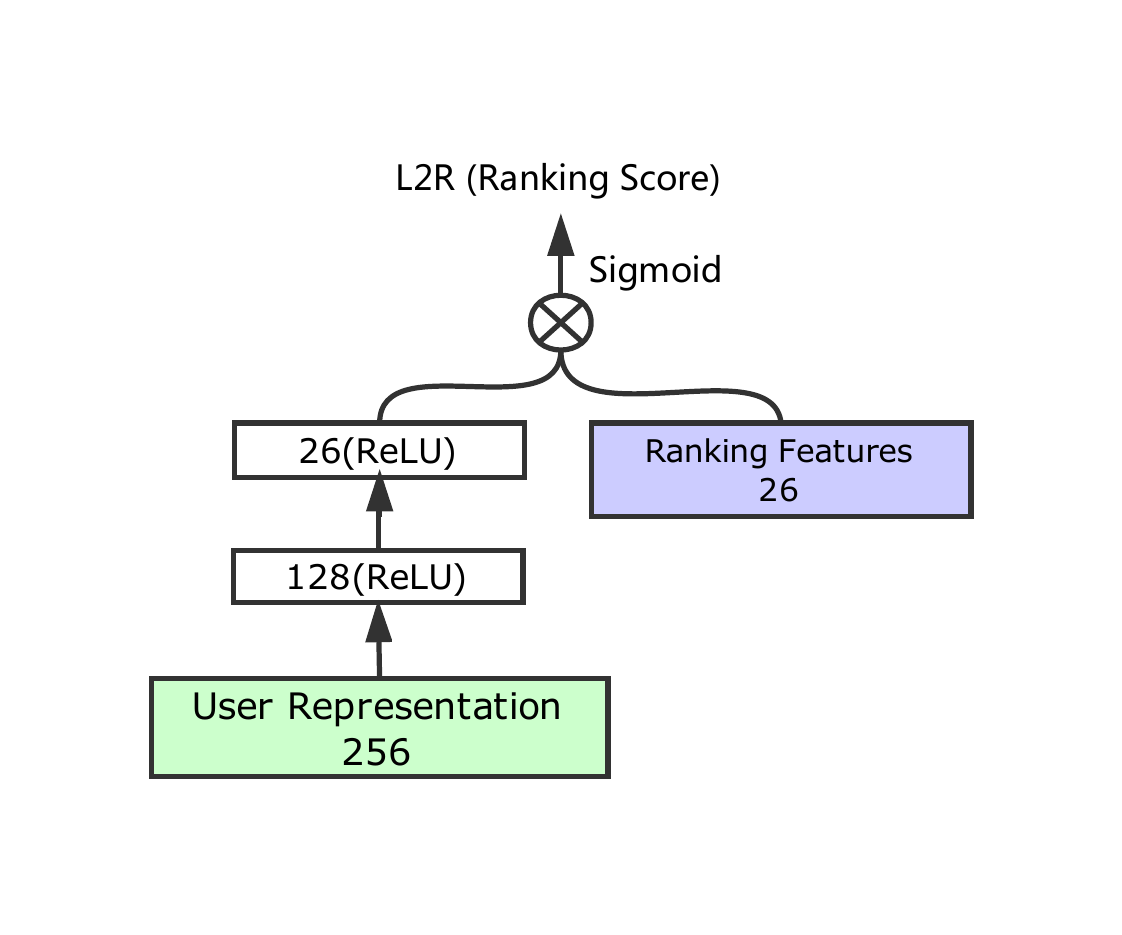}
		\end{minipage}
	}
	\subfigure[PPP task, only depend on the user representation] {
		\begin{minipage}[b]{0.175\textwidth}
			\includegraphics[width=0.85\textwidth]{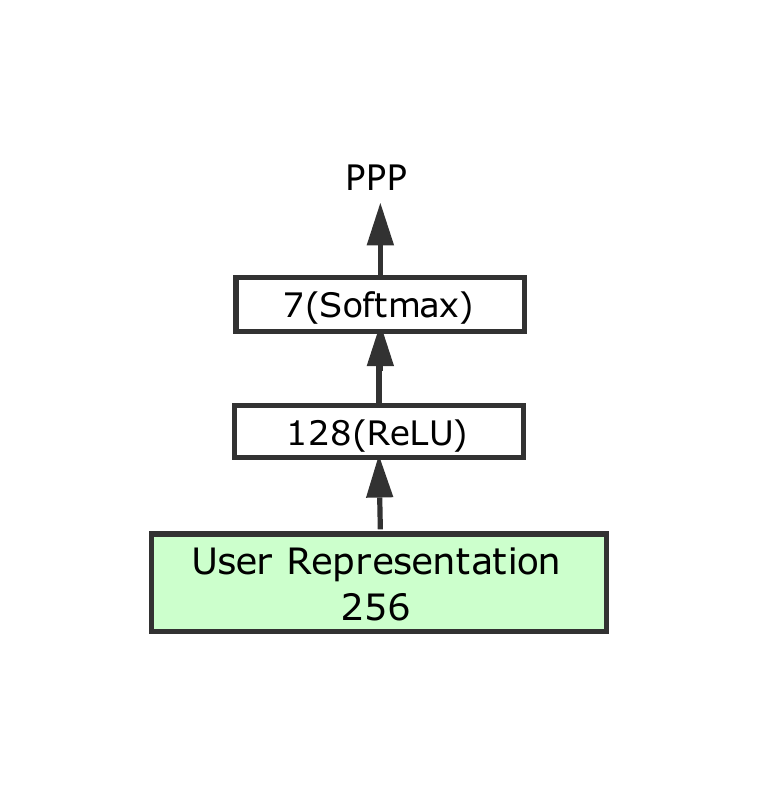}
		\end{minipage}
	}
	\subfigure[FIFP task, a diverse task, reflecting users relations]{
	\begin{minipage}[b]{0.19\textwidth}
		\includegraphics[width=0.985\textwidth]{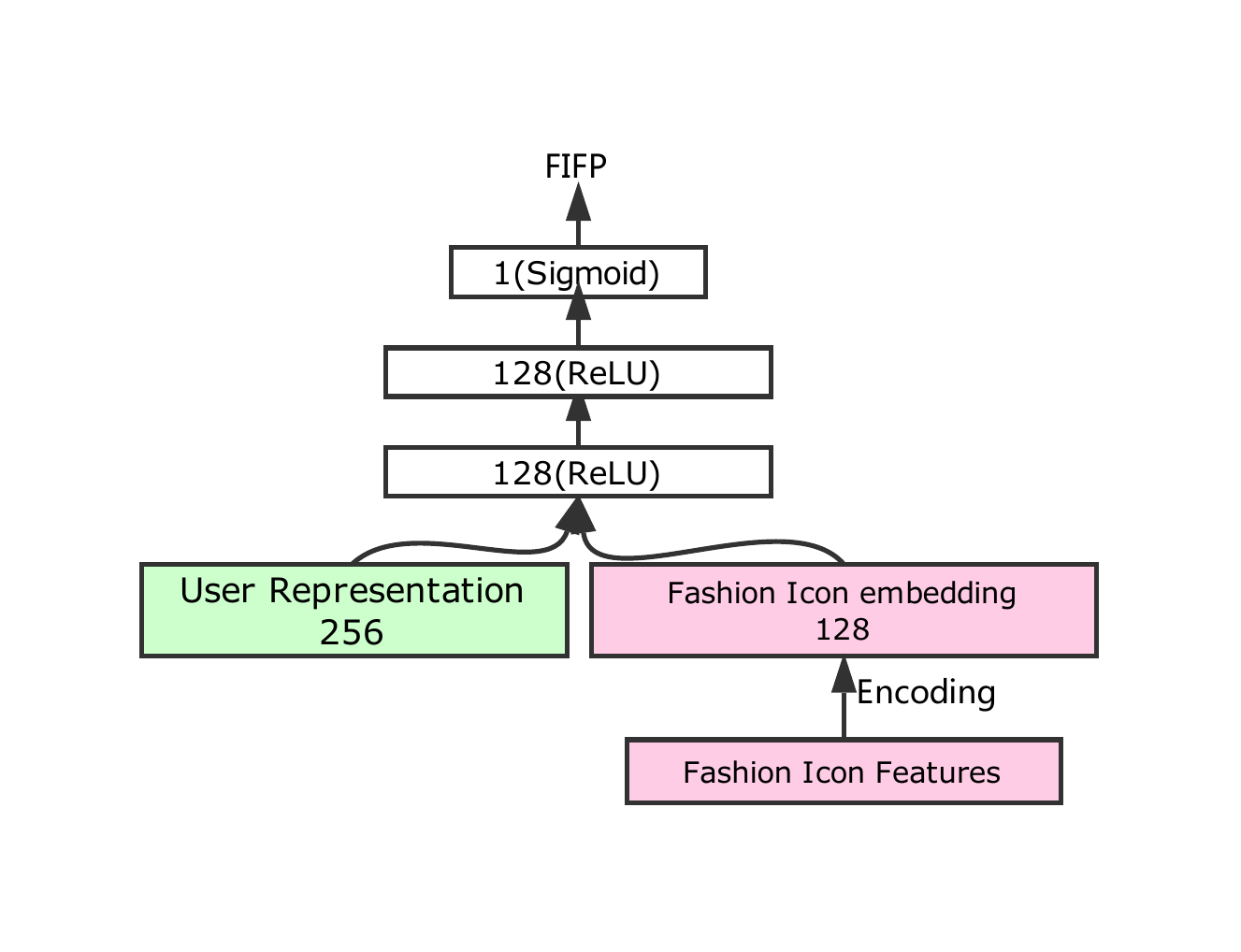}
	\end{minipage}
	}
	\subfigure[SPP, a task for validating the transferability]{
	\begin{minipage}[b]{0.19\textwidth}
		\includegraphics[width=0.985\textwidth]{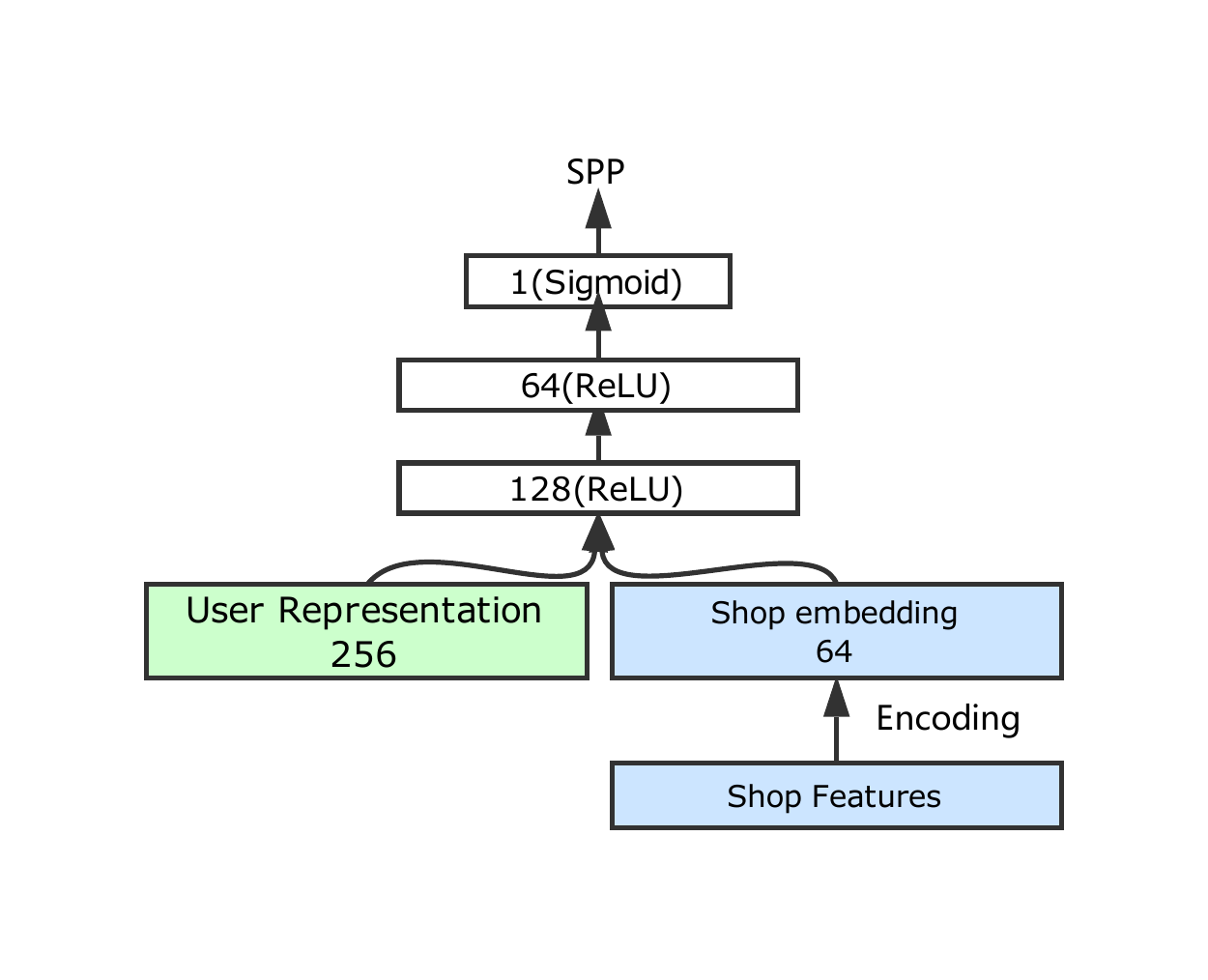}
	\end{minipage}
	}
	\caption{Net architecture of different tasks, user representation is shared by different tasks. 
}                     
	\label{figmuti-task}   
\end{figure*}

	 This paper is aiming to generate universal user representations. As shown in Figure ~\ref{fig:Model}, after obtaining the user representation, we define several related tasks to learn simultaneously. For each task, the other tasks are viewed as regularizations. By sharing representations with multi-task learning, we can enable the user representations general and reliable. We define five tasks illustrated in Figure ~\ref{figmuti-task}. Though all the tasks are trained offline, they should make realtime predictions online, as user behaviors change over time.

\par

	\textbf{Click Through Rate Prediction(CTR):} CTR task takes user representation $rep_i$ and current item representation $e_i$ as input, and aims to learn the probability that user $rep_i$ clicks item $e_i$. The predicted item CTR can be used as a personalized feature for ranking. CTR prediction is a classification task and the loss function is the likelihood defined as follows:

	\begin{equation}\label{eqctr}
	\begin{split}
	\mathscr{L}_{CTR}(\alpha) = &-\frac{1}{N}\sum_{i=1}^N [ y_ilog(score(rep_i,e_i;\alpha))\\
&+(1-y_i)log(1-score(rep_i,e_i;\alpha)]
	\end{split}
	\end{equation}
	where $rep_i$ denotes the user representation of the $i^{th}$ sample which is the output of the attention net, $e_i$ denotes the item representation,and $y_i$ denotes the label of the $i^{th}$ sample, which is set to 1 if user $i$ clicked item and 0 otherwise. $score(rep_i,e_i)$ is a function that maps $rep_i$ and $e_i$ to a real valued score, indicating the probability whether the user will click the item. We implement $score(\cdot;\alpha)$ with another shallow neural network as illustrated in Figure ~\ref{figmuti-task}(a).
\par
\textbf{Learning to Rank(L2R):} Personalized L2R task takes user representations and ranking features as input, and aims to learn the weights of the ranking features to maximize the conversion rate. We use point-wise L2R in our work. The loss function is defined as Equation ~\ref{eqltr}:
	\begin{equation}\label{eqltr}
	\mathscr{L}_{L2R}(\theta) = \sum_in_ilog(1+exp(-y_iweight( rep_i;\theta )^Tr_i))
	\end{equation}

	where $y_i\in \{-1,+1\}$ is the label of $i$th sample, $n_i\in R$ indicates the sample weight which is different according to different user behaviors types. Generally, the weights are pre-defined by some business rules, e.g. the weight of a purchase instance is usually higher than a click instance. $r_i\in R^m$ is the ranking features (some are listed in Table 1),  $weight(rep_i;\theta )$ is a function which maps $rep_i$ to a  $m$-dimensional vector which indicates the weights of ranking features. The net architecture is illustrated in Figure ~\ref{figmuti-task}(b).
\par
\textbf{Price Preference Prediction(PPP):} We treat user price preference prediction as a multi-class classification task.  $p = \{p_1, p_2, \ldots, p_k\}$  denotes the $k$ classes of item price, each $p_i$ indicates a price range. $p_1$ is the cheapest range and $p_k$ is the most expensive range. This task is actually predicting the price range of item that the user is going to purchase. The loss function is defined as Equation  ~\ref{eqp}:  
	\begin{equation}\label{eqp}
	\begin{split}
	&\mathscr{L}_{Price}(\sigma) = -log(\sigma_y(z)) \\
	&s.t. \sigma_i(z) = \frac{exp(z_i)}{\sum_{j=1}^kexp(z_j)}, i=1,\ldots,k \\
	&z_i=predict(rep,i;\gamma)
	\end{split}
	\end{equation}
	where $y\in \{1,2,\ldots,k\}$ is the label, $z_i$ is the predicted score of class $i$, $z_i$ takes user representation $rep$ as input, $predict(rep,i;\gamma)$ is a function maps user representation $rep$ to a real valued score and $\sigma_i(z)$ indicates the probability that the user prefers price range $i$. In our work, the item price is divided into 7 levels.
\par
	\textbf{Fashion Icon Following Prediction(FIFP):} In Taobao, there are a lot of fashion icons who provide fashion suggestion for wearing, making up, etc. Common Taobao users may follow some of them. FIFP is somehow a diverse task reflecting user relations. It takes user representation $rep_i$ and features of the fashion icon $f_i$ as input, and aiming to learn the probability whether user $rep_i$ will follow  $f_i$. FIFP is also a classification task and the net architecture is shown in Figure ~\ref{figmuti-task}(d).
\par
	\textbf{Shop Preference Prediction (SPP):} Taobao is an e-commerce portal 
	on which billions of sellers operate their own online shops. In this task, we predict the preferred shops of each user. The task is not simultaneously trained as one of the multiple tasks, but is treated as a transfer task for the well learned DUPN model and user representation.
	
\section{Experimental Methodology}
\label{sec::Exp_Meth}

	\textbf{Dataset Description:} Experiments are conducted on a large-scale offline benchmark dataset. The overall offline benchmark dataset consists of 5 subsets corresponding to 5 different tasks. About $6\times10^9$ instances are drawn from the daily log files of different scenarios in Taobao, describing the user historical behaviors and the label of each task. We use samples across 10 days for training and evaluate on the data of the next day. Instances in the large-scale data set are shuffled and split into smaller batches, each of which includes 1024 instances.
	
	\par
	\textbf{Online Environment:} We deployed the algorithm in the operational Taobao search system. Two tasks of DUPN will directly affect the ranking performance. First, the task of CTR produces an estimate click probability as a feature for ranking. Second, the task of L2R combines all the ranking features and produces a final ranking score. Besides, the task of PPP is also applied for user analyzing. The three tasks are predicted simultaneously in real time.
	
	There are more than 100 million unique users searching for products in Taobao search each day, from which 10\% are randomly selected as the experiment group for A/B testing.

	\par
	\textbf{Experimental Configuration:} We set the hyper-parameter configurations as follows. The length of the behavior sequences is set to 100. We apply dropout~\cite{srivastava2014dropout} to the output of all fully connected layers and the dropout rate is set to 0.8. L2-regularization is also applied to prevent the neural networks from overfitting. The proposed network is trained with SGD, using AdaGrad~\cite{duchi2011adaptive} and a learning rate of $10^{-3}$.

	We train DUPN utilizing a distributed TensorFlow~\cite{abadi2016tensorflow} machine learning system, using 96 parameter servers and 2000 workers, each of which runs with 15 CPU cores. The system can train 400 batches per second and the total training process takes more than 4 days.

\section{Experiment and Analysis}
	In this section, we compare DUPN with some competitive models proposed in recent literatures. First we demonstrate advantages of the proposed network architecture on our learning tasks. Then, the transferability and generalization of the learned user representations are verified. Finally, we give the performance and case study in the operational environment.

\subsection{Offline Comparison of Different Networks} 

	We use the following representative state-of-the-art methods on our benchmark dataset to highlight the effectiveness of DUPN. In these experiments, the tasks are learned independently.
	
	\textbf{Wide:} A linear logistic regression model widely used in large-scale search systems. User profiles, item features and rich cross-product feature transformations are the model input.  We use the Follow the-regularized-leader (FTRL ~\cite{mcmahan2011follow}) as the optimizer.
	
	\textbf{Wide \& Deep~\cite{cheng2016wide}:} A model combines a logistic regression model and a deep neural network together, to maintain the benefits of memorization and generalization. AdaGrad and FTRL are used as optimizers in deep side and wide side separately. In the deep side, the sizes of the 3 hidden layers are 1024, 512 and 128.
	
	\textbf{DSSM~\cite{huang2013}:} The model was proposed for representing the text strings. In this study, we apply DSSM  to model the similarity of the user-query and the item. In each instance, the user-query are encoded by 3 hidden layers with the size of 1024, 512 and 128, the items  are encoded with the same net structure. Then the preference are calculated by the cosine similarity.
	
	\textbf{CNN-max~\cite{zheng2017joint}:} A model uses CNN with max pooling to encode the user behavior history. We apply window sizes from one to ten to extract different features, and all of the feature map have the same kernel size of 64. A max pooling layer is added on the top of feature map, the output is then passed to a fully connected layer to produce final user embedding.
	
	\textbf{DUPN-nobp/bplstm/bpatt:} Sub-models of DUPN, in which the behavior property is not used, only used in the gates of LSTM and only used in the attention net respectively.
	
	\textbf{DUPN-all} :The full proposed model in this paper.	
	
	\textbf{DUPN-w2v} :a sub-model of DUPN, in which the item embeddings are pre-trained by word2vec ~\cite{Mikolov2013Efficient}, instead of in an end-to-end manner. In the pre-training procedure, we treat each user's behavior sequence as a sentence, and the sentences from all the user compose the word2vec learning document.

	We evaluate the performance of different models for the multiple tasks separately. For CTR, L2R, FIFP, and SPP tasks, Area Under the Curve (AUC) is used to investigate the effectiveness. And we use precision as the evaluation metric for the task of PPP. The comparison results are presented in Table ~\ref{different deep network methods}.	\par
	
	The first 4 rows of Table ~\ref{different deep network methods} report the results of some state-of-the-art baselines. Of all the baselines, Wide \& Deep model and CNN-max model are shown to be competitive. Rows 5-8 show the results of DUPN and some of its sub-models. Performance of DUPN-nobp is similar to CNN-max. DUPN-bplstm and DUPN-bpatt perform much better than the baselines and DUPN-nobp, the behavior properties prove to be effective, when they are used both in the gates of LSTM and in the attention mechanism.  DUPN-all outperforms all the other models.
	
	The last row reports the result of DUPN with pre-training. DUPN-all outperforms DUPN-w2v about 2.0\% in terms of AUC of L2R, 2.4\%  of CTR, 3.6\% of FIPP and 6.0\% in terms of precision of PPP. The precision improvement is especially outstanding and we prove that end-to-end learning can improve the model training when we have a large enough training set. Item embeddings pre-trained by word2vec mainly depend on the item co-occurrence, so it can only learn the similarity of the items. In contrast, DUPN-all can extract more distinct information of the items such as popularity.

	\begin{figure*}[htbp]
	\centering
	\subfigure[the comparison of L2R task]{
		\begin{minipage}[b]{0.21\textwidth}
			\includegraphics[width=0.85\textwidth,height=3cm]{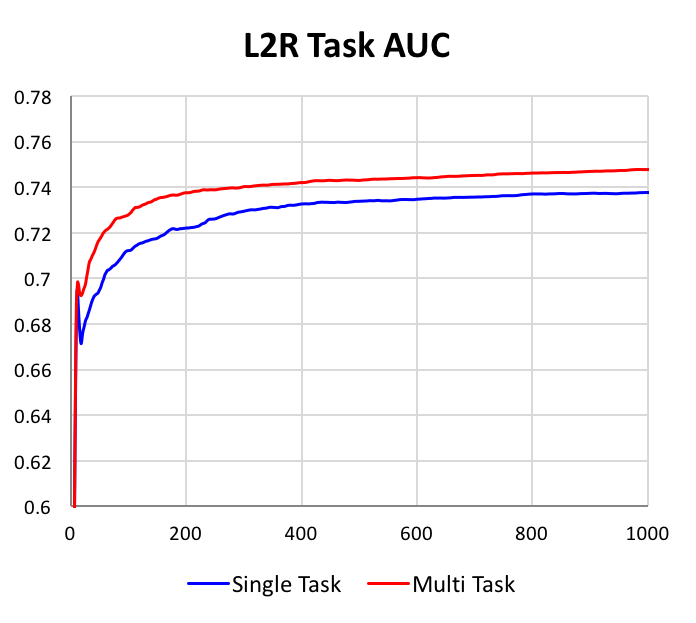} \\
			\includegraphics[width=0.85\textwidth,height=3cm]{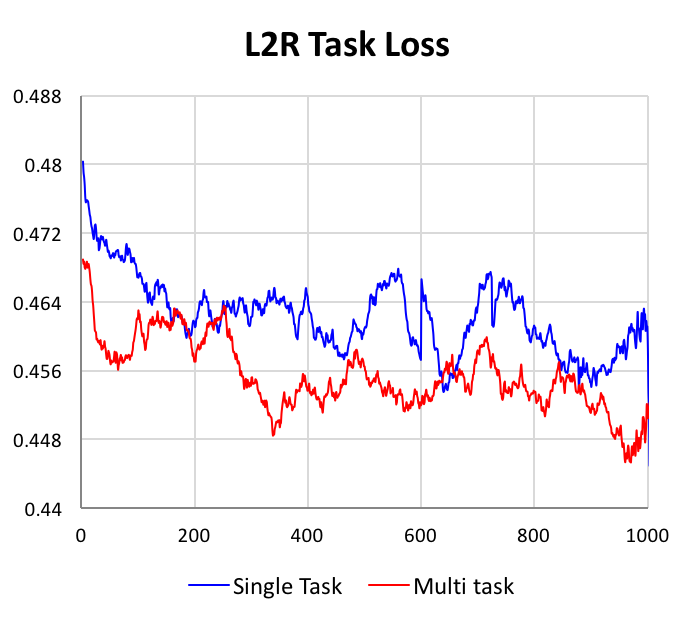}
		\end{minipage}
	}
	\subfigure[the comparison of CTR task]{
		\begin{minipage}[b]{0.21\textwidth}
			\includegraphics[width=0.85\textwidth,height=3cm]{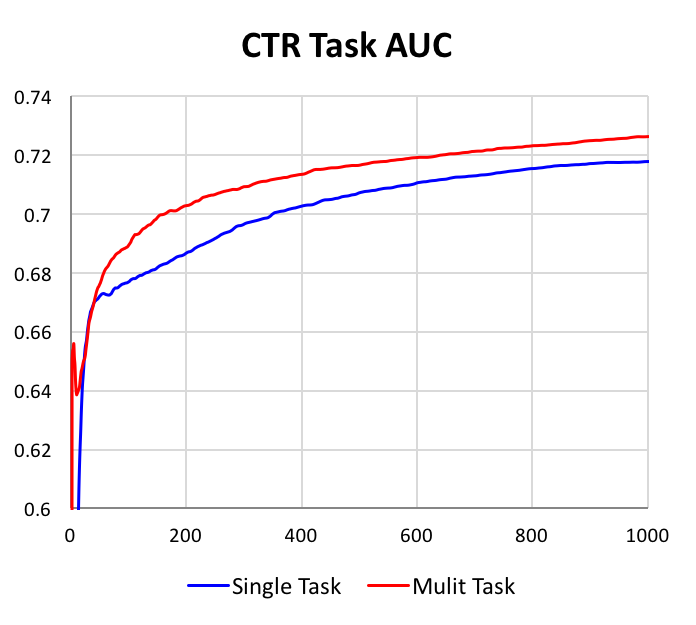} \\
			\includegraphics[width=0.85\textwidth,height=3cm]{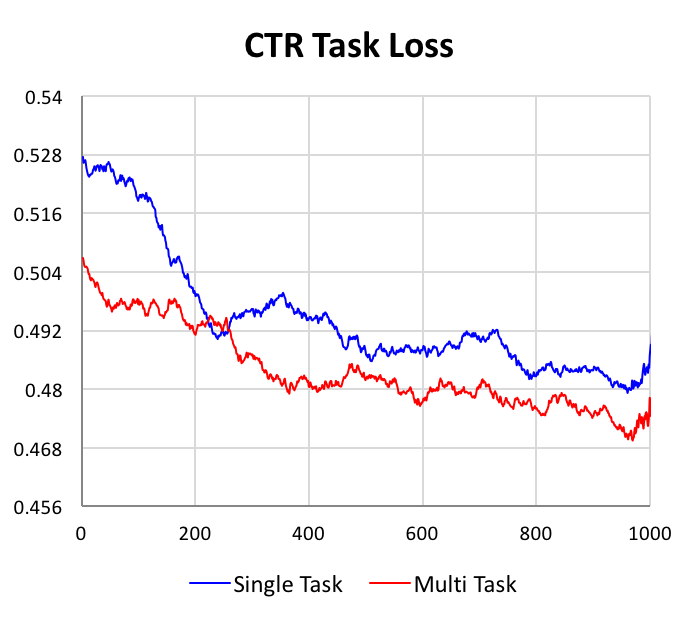}
		\end{minipage}
	}
	\subfigure[the comparison of PPP task]{
		\begin{minipage}[b]{0.21\textwidth}
			\includegraphics[width=0.85\textwidth,height=3cm]{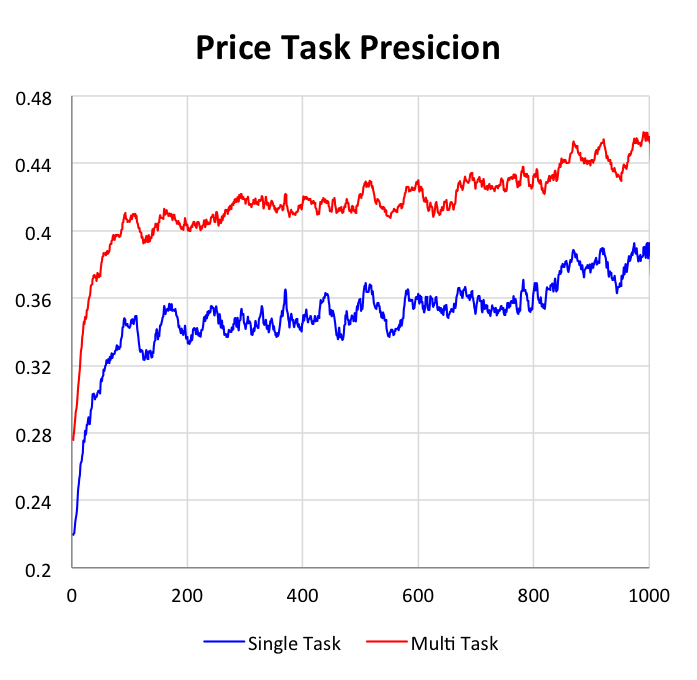} \\
			\includegraphics[width=0.85\textwidth,height=3cm]{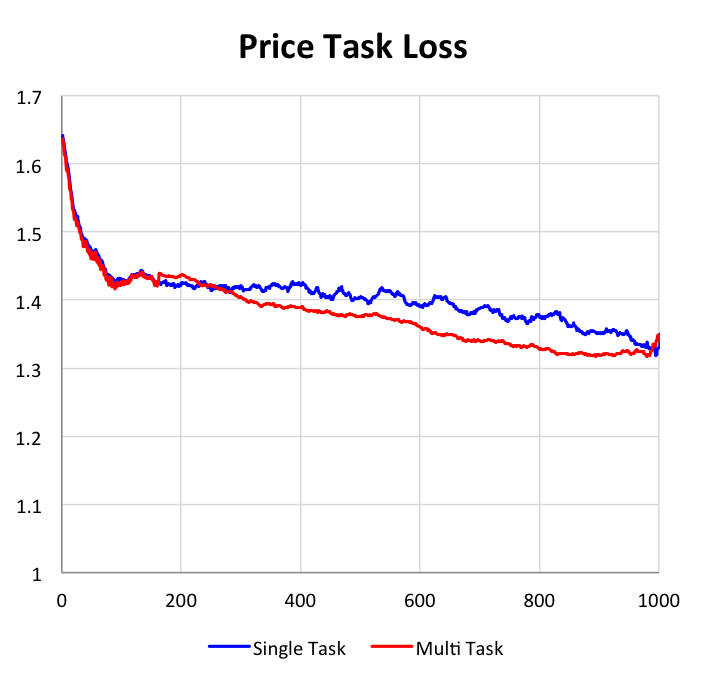}
		\end{minipage}
	}
	\subfigure[the comparison of FIFP task]{
	\begin{minipage}[b]{0.21\textwidth}
		\includegraphics[width=0.85\textwidth,height=3cm]{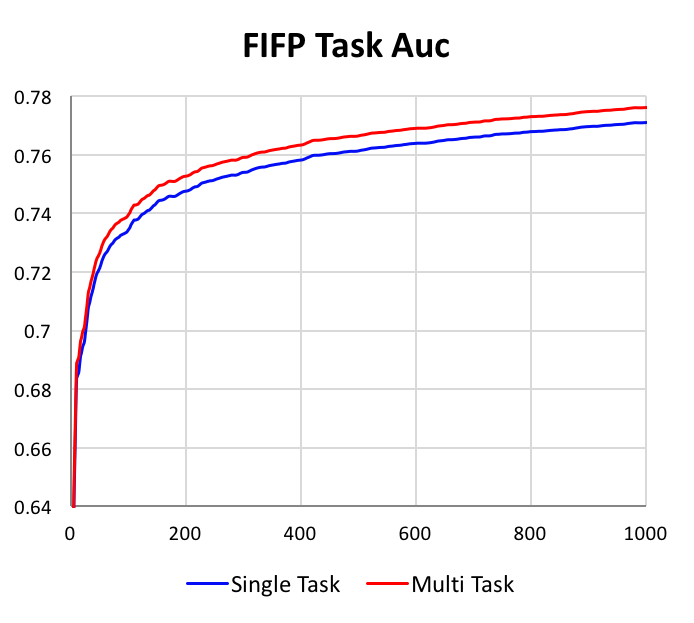} \\
		\includegraphics[width=0.85\textwidth,height=3cm]{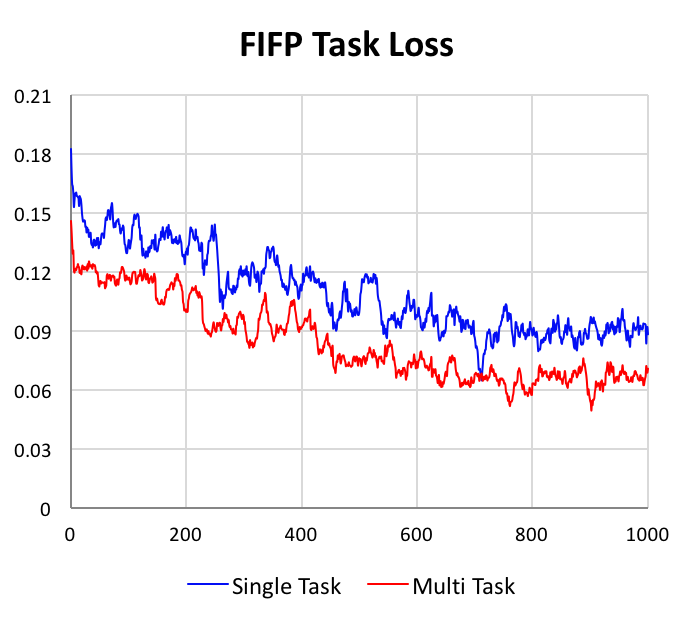}
	\end{minipage}
	}
	\caption{Comparison of single task and multi-task on L2R, CTR PPP and FIFP task respectively. x-coordinate stands for the learning iterations with space interval 100000 and y-coordinate presents the evaluation metric. The curve of AUC and Precision is smoothed by the contiguous batch points. 
}                     
	\label{figmuti}   
\end{figure*}

\begin{table}[t]
	\centering
	\caption{Comparison of Different Models}
	\begin{tabular}{c|cccc}
		\hline
		 & L2R & CTR & FIFP & PPP \\
		 & AUC  & AUC & AUC & Precision \\
		\hline
		Wide & 0.70502 & 0.68266 & 0.71839 & 34.094\% \\
		
		Wide \& Deep & 0.71581 & 0.69957 & 0.74581 & 38.581\% \\
		
		DSSM &  - & 0.68961 & 0.72035 & - \\
		
		CNN-max & 0.72391 & 0.70735 & 0.73803 & 39.904\% \\
		\hline 
		DUPN-nobp & 0.73307 & 0.70221 & 0.74082 & 39.394\% \\
		
		DUPN-bplstm & 0.74583 & 0.72139 & 0.75337 & 42.926\% \\
		
		DUPN-bpatt & 0.73901 & 0.71303 & 0.75927 & 41.350\% \\
		
		DUPN-all & \textbf{0.75005*} &   \textbf{0.72519*} &   \textbf{0.77323*} & \textbf{44.011\%*} \\
		\hline 
		DUPN-w2v  & 0.73091 & 0.70127 & 0.73749 & 38.079\% \\
		\hline
	\end{tabular}
	\label{different deep network methods}
\end{table}

\subsection{Single Task VS Multi-task} 

	Each task can be learned independently or together with the other three tasks. By sharing shallow features and representations between related tasks, we can enable our model to generalize better for all learning tasks. We prove the above benefits by comparing single task learning with multi-task learning.  

	We experiment on the benchmark dataset with four single tasks. As shown in Figure ~\ref{figmuti}, the learning process can be divided into 2 periods. In the first period the AUC increases rapidly while in the later period the AUC grows slowly and persistently for several days. This is mainly because of the imbalance of the features. Some of the features are frequent such as item tags while some other features are very sparse such as item IDs. Frequent features can be learned quickly as they appear in most of training instances but sparse features need a much longer learning period as they are included in less instances. However sparse the features are, they all contribute to the high AUC or precision of DUPN.

	A more important thing we observed is that the multi-task learning performance outperforms single task learning in all the tasks. Sub-figures(a)(b) and (d) show the performance of L2R, CTR and FIFP tasks trained with single task and multi-task learning respectively. Multi-task learning converges faster from the very beginning  and gains about 1\% improvement than single task finally. The precision of price preference task increases from 40\% to 44\% shown in sub-figure(c) and it is a statistically significant improvement over other baselines. The 4 comparisons demonstrate that multi-task learning enables the model to extract more useful information and learn general representation that are helpful for each task. 
	
	Though the improvements of some tasks are not so significant, one model for multiple tasks are much more applicable than several independent models. For a realtime online system such as ranking, some of the tasks (e.g., CTR CVR and PPP) should be inferred simultaneously, so a nonnegligible advantage is that one model takes much less CPU cost, memory and response latency.
	
\subsection{Representation Transferability} 

	DUPN is trained to obtain universal user representations. In this subsection we verify that if the user representations could be easily transferred into new related tasks. 
	
	Supposing that DUPN with 4 tasks has already been applied online, and we want to further predict the shop preference of the users (described in Section 4.3). Generally we have  four choices as follows:
	
	\textbf{End-to-end Re-training with Single task (RS)}. Shop preference is treated as a single task using the architecture of DUPN. The sophisticated model is retrained independently and the learning process is the same as that in former 4 tasks.
	
	\textbf{End-to-end Re-training with All tasks (RA)}. DUPN model is retrained for shop preference together with the other 4 tasks. Through the other 4 tasks are learned together, they are auxiliary tasks helping to learn general features.
	
	\textbf{Representation Transfer (RT)}. This method directly uses the learned user representations from the existing DUPN, instead of employing a large-scale and complex new network. A simple and shallow network is used for the classification task, where the user representations and the shop features are the only inputs.
	
	\textbf{Network Fine Tuning (FT)}. A shallow network for the new task is added on the representation layer of DUPN. The learned model of DUPN is used as an initialization, and the whole network is fine tuned for the new task.
	
	\begin{figure}[t]
	\centering
	\includegraphics[width=1.0\linewidth]{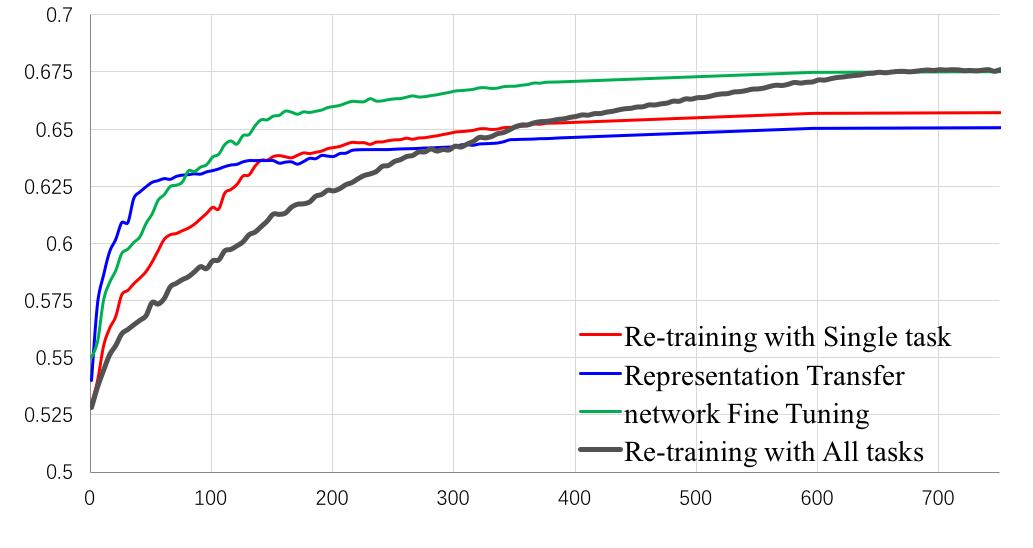}
	\caption{Comparison of 4 methods of training a related new task, i.e., shop preference. x-coordinate stands for the learning iterations with space interval 100000 and y-coordinate presents the evaluation metric of AUC. }
	\label{fig:Transfer}
\end{figure}

	Figure ~\ref{fig:Transfer} illustrates the results of the four methods. Firstly, we can observe that FT and RA both achieve the best result while FT has a higher rate of convergence. The corresponding AUC is 2.5\% higher than RS and 3\% higher than RT. This comparison proves that the existing DUPN model can well extract general features from user behavior sequence. Fine tuning is a appropriate method for a new related task and any kind of retraining is not necessary. However, the problem is that the new task introduce a new big model, which will bring more online memory and CPU cost, and higher response latency. 
	
	To prevent the online search service from too many complex deep models, RT is another proper method for the new tasks. The rate of convergence of RT is the highest because of its smallest parameter space.  As training process goes on, final AUC of RT is exceeded by other training methods, however the gap is only 0.5\% compared to RS and 3\% compared to FT and RA. In spite of the slightly lower AUC, RT may be more welcome for an online system due to its simplicity and higher inference efficiency. The comparison shows that the user representation can be directly used in related tasks such as shop preference, achieving a higher rate of convergence and a fine performance. The results also demonstrate the generalization capability and transferability of the user representation.

\subsection{Case Study of Attention} 

	In DUPN, attention-based pooling is used to determine the importance of different items in the user behavior sequence under a certain query and user. Section 6.2 illustrated the effectiveness of attention mechanism and in this subsection we further give some presentational cases on how attention pooling works in the e-commerce scenario.
     	
\begin{figure}[t]
	\centering
	\includegraphics[width=1.0\linewidth]{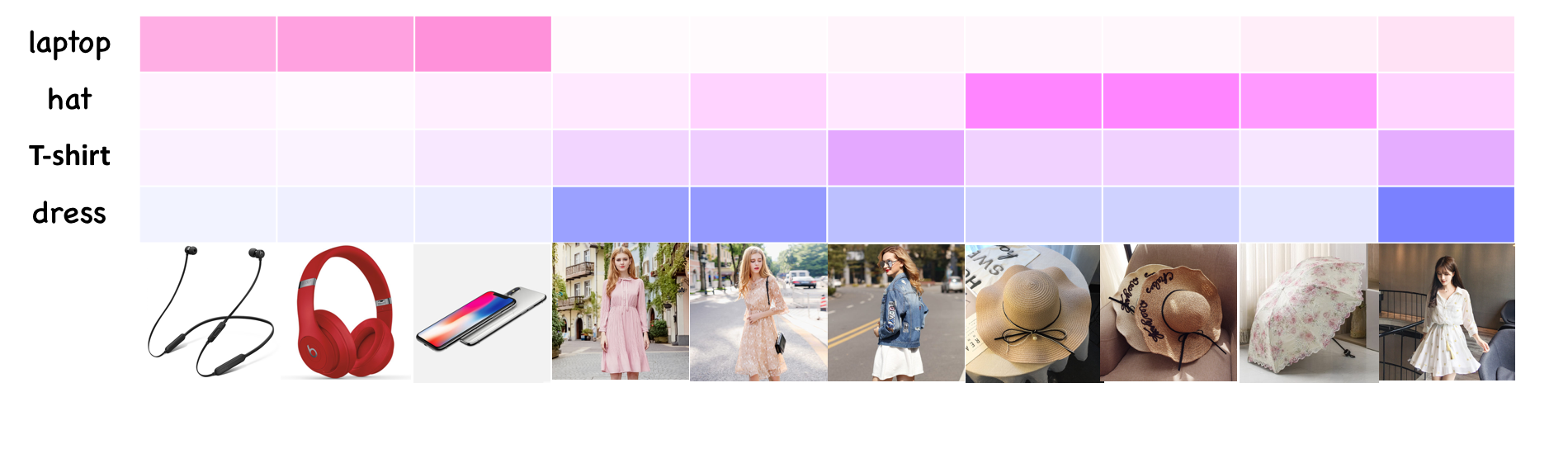}
	\caption{Examples of attention weights with different queries to a specific user behavior sequence. The higher attention weights of the item, the darker color of the grid.}
	\label{fig:Attention}
\end{figure}

	We first randomly select an online customer of Taobao to study the query impact of attention-based pooling. The attention weights of different items in the selected behavior sequence under different query are visualized in Figure ~\ref{fig:Attention}. The last row denotes the items sequence ordered by behavior time, to the right are new clicks. Each one of the first 4 rows is a group of attention weight assignment under a certain query. As we can see, when the user searches "dress", we find that attention layer highlights weights of dress than jacket or hat while electronic products get rare attention. Similarly, the historical items of hat gain more importance than other items when searching "hat", and when we issue the query "laptop", earphone and cellphone dominate most. The above cases illustrate that the attention mechanism in DUPN can successfully extract information related to the context while eliminate irrelevant items. 
	
\begin{figure}[t]
	\centering
	\includegraphics[width=1.0\linewidth]{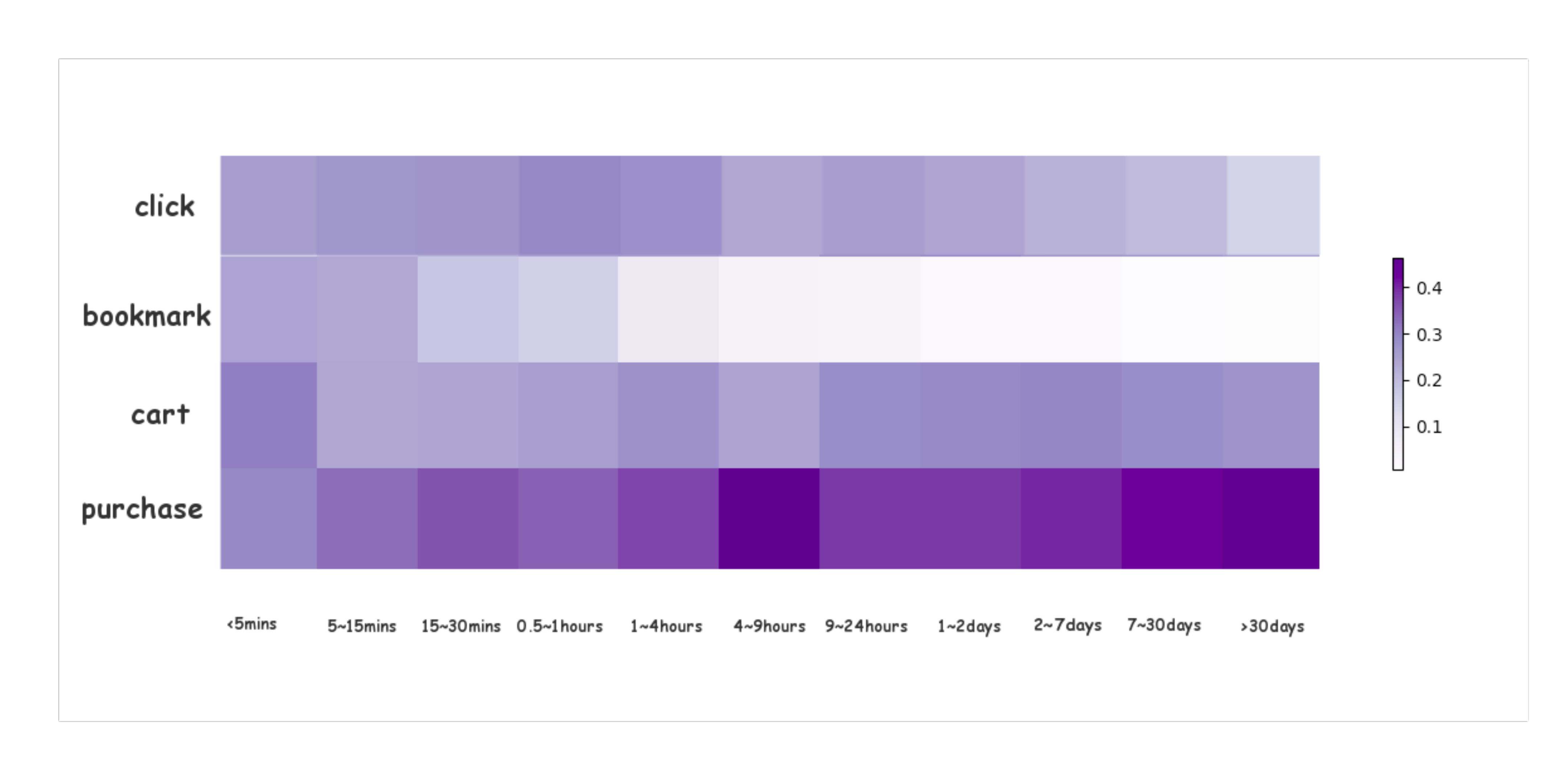}
	\caption{Heat map of of attention weights with different behavior types of different time.}
	\label{fig:type_time}

\end{figure}

	We further study how the behavior properties affect the attention mechanism. Properties of behavior type and time are intensively examined. Behavior types include click, bookmark, add to cart and purchase, and behavior time are split into several buckets. Heat map matrix of Figure ~\ref{fig:type_time} represents the attention scores of each behavior type-time pair. We can have several observations of this figure. Firstly, different types of behaviors contribute differently in generating user representation. Purchase behavior is generally the most important since the colors of its heat distribution is overall darker than other behavior types. Behaviors of click and add to cart are less important and behaviors of bookmark take least attention. Secondly, attention scores are greatly affected by the behavior time. Recent clicks take more attention than earlier clicks, because user interest may change over time and recent behaviors can represent her(his) latest interest. Same regularity can be found in behaviors of bookmark. However, very recent purchase behaviors seem to be less crucial than older ones. This is because purchase usually means one's shopping demand is temporarily satisfied.
Heat distribution of cart seems similar to purchase, except that the cart behaviors within 5 minutes are still very important because in this time period, products in the cart usually have not been bought.

\subsection{Online A/B Testing} 

\label{sec::d11}

	This subsection presents the results of online evaluation in Taobao search system with a standard A/B testing configuration. We concern most on two important commercial metrics, i.e. CTR and sales volume. Besides, we also compare the precision rate of price preference prediction on DUPN with former online model based on MaxEnt~\cite{Berger1996A}. 
	
\begin{figure}[tbp]
	\centering
	\subfigure[Recall rate of Price Level]{
		\begin{minipage}[b]{0.22\textwidth}
			\includegraphics[width=0.98\textwidth]{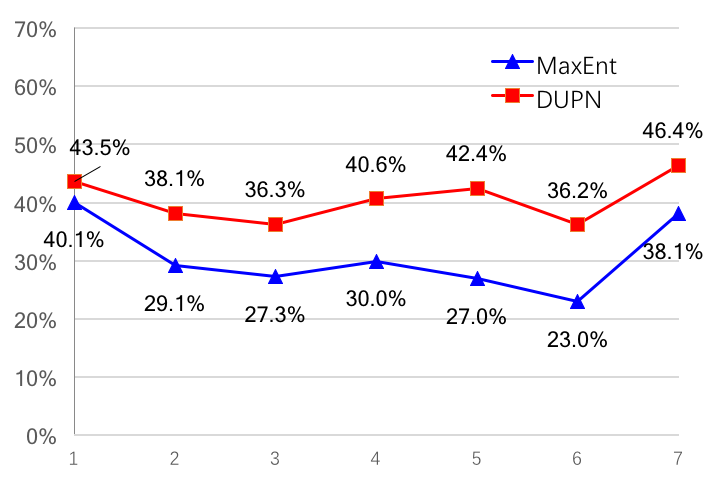} 
		\end{minipage}
	}
	\subfigure[Precision of Price Level]{
		\begin{minipage}[b]{0.22\textwidth}
			\includegraphics[width=0.98\textwidth]{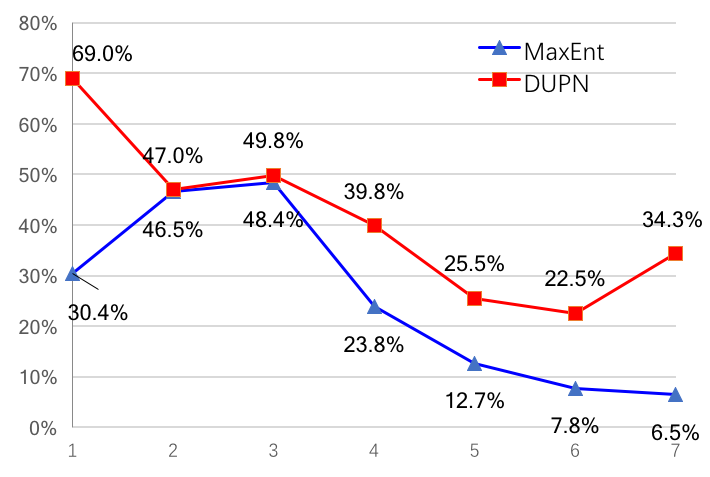}
		\end{minipage}
	}
	\caption{Online effect comparison of price preference task. It is a 7-label classification problem, 2 sub-figures report the recall rate and precision on each of the labels.}                     
	\label{fig:Price}
\end{figure}

	Table 3 reports the results of improvement of CTR, sales volume and price precision rate when we apply DUPN on the operational search system. The results show that CTR can increase 2.23\% and sales volume can increase 3.17\% in an average of 7 days. The overall price precision can also increase from 33.2\% to 44.2\%. Figure ~\ref{fig:Price} presents a more detailed result of the precision and recall on each price class. Recall rate of 7 classes is balanced and recall of DUPN is 2\% to 10\% higher than MaxEnt. Precision of different classes is more diverse, while DUPN still outperforms MaxEnt in all classes, especially for the cheapest items and the most expensive items.

\begin{table}[t]

	\centering
	\caption{Online A/B Testing}

\begin{center}
\begin{tabular}{ c|ccc} 
\hline

\multirow{2}{2em}{Day} & CTR & Sale& Price\\
&  Improve &  Improve & Precision\\ 
\hline
1
& 2.32\% & 2.93\% &34.4\% $\to$ 43.5\%\\ 
\hline
2
& 2.17\% & 3.15\% &32.6\% $\to$ 44.2\%\\ 
\hline
3
& 2.13\% & 3.26\% &32.9\% $\to$ 44.7\%\\ 
\hline
4
& 2.21\% & 3.07\% &33.3\% $\to$ 45.1\%\\ 
\hline
5
& 2.31\% & 3.23\% &33.0\% $\to$ 43.7\%\\ 
\hline
6
& 2.27\% & 3.37\% &32.3\% $\to$ 43.9\%\\ 
\hline
7
& 2.19\% & 3.19\% &34.8\% $\to$ 44.6\%\\ 
\hline
Average
& \textbf{2.23\%} & \textbf{3.17\%} & \textbf{33.2\% $\to$ 44.2\%}\\ 
\hline
\end{tabular}
\end{center}
\label{Online A/B Test}
\end{table}

\section{Practical Guidelines on Model Implementation}
This section provides some practical guidelines on large scale neural network model implementation in Taobao search system.

\textbf{Daily incremental updating:} Because user preference may change over time, and items, brands or styles go in and out of fashion, frequent model updating is necessary. However, although the model is trained in a distributed TensorFlow system of 2000 workers, each of which contained 15 CPU cores, it still takes three or four days to train data of 10 days.
So, we adopt the method of incremental learning to solve this problem, where the model is trained with data of 10 days only for the first time, and is fine-tuned with new data day by day later. By doing this, the training time can decrease to less than 10 hours and the model can fit better in the online environment. By incremental updating, the model can be fine-tuned everyday with latest online data, while by  learning only once, new data and old data are treated equally.

\textbf{Network disassembly:} When a user issues a query, Taobao search system should calculate scores for about ten thousand of items, meaning that the large scale network should be inferred ten thousand times. This is unacceptable for a real-time system with high demand of computing efficiency. 

An additional advantage of DUPN is that the network architecture is splittable. As user features and item features are not crossed in the shallow layers, we can divide the integrated network into two parts. The first part includes generating of user representations and tasks only based on the representations, such as PPP. This part of the network takes most of the calculations. The second part only includes tasks based on both user and item representations. Network structure of this part is simple and has much smaller computational requirements.
Under a certain query, the input of the proposed network on the first part is fixed, denoting that it may infer only once.
After obtaining the user representations from the first part, shallow network on the second part forwards thousands of times for CTR prediction and ranking scores of the items.

\section{Conclusion}
This paper proposes a robust and practical user representation learning algorithm in e-commerce named DUPN. LSTM and attention mechanism help to model heterogeneous behavior sequence, and behavior properties facilitate valid LSTM memorizing and attention weights allocation. Benefiting from better information utilization of multiple tasks, the user representations are more effective to reflect their interests. We further provide some practical lessons of learning and deploying the large operational deep learning model in an operational e-commerce search system.
\par
An extensive set of experiments are provided to show the competitive performance of DUPN and generality and transferability of the user representation. Detailed discussion of some case studies are also provided to show the insight of the how the attention mechanism works.
DNPN has already been deployed on the online e-commerce search system of Taobao. Online A/B testing results show that our model meets user's preference better and improves customer's shopping efficiency.

\section{Acknowledgement}
We thank colleagues of our team - Zhirong Wang, Xiaoyi Zeng, Ling Yu and Bo Wang for valuable discussions and suggestions on this work. We thank our search engineering team for the large scale distributed machine learning platform of both training and serving. We also thank scholars of prior works on representation learning and recommender system. We  finally thank the anonymous reviewers for their valuable feedback.


\bibliographystyle{ACM-Reference-Format}
\balance
\bibliography{ibrain}


\begin{thebibliography}{00}


\ifx \showCODEN    \undefined \def \showCODEN     #1{\unskip}     \fi
\ifx \showDOI      \undefined \def \showDOI       #1{#1}\fi
\ifx \showISBNx    \undefined \def \showISBNx     #1{\unskip}     \fi
\ifx \showISBNxiii \undefined \def \showISBNxiii  #1{\unskip}     \fi
\ifx \showISSN     \undefined \def \showISSN      #1{\unskip}     \fi
\ifx \showLCCN     \undefined \def \showLCCN      #1{\unskip}     \fi
\ifx \shownote     \undefined \def \shownote      #1{#1}          \fi
\ifx \showarticletitle \undefined \def \showarticletitle #1{#1}   \fi
\ifx \showURL      \undefined \def \showURL       {\relax}        \fi
\providecommand\bibfield[2]{#2}
\providecommand\bibinfo[2]{#2}
\providecommand\natexlab[1]{#1}
\providecommand\showeprint[2][]{arXiv:#2}

\bibitem[\protect\citeauthoryear{Abadi, Agarwal, Barham, Brevdo, Chen, Citro,
  Corrado, Davis, Dean, Devin, et~al\mbox{.}}{Abadi et~al\mbox{.}}{2016}]%
        {abadi2016tensorflow}
\bibfield{author}{\bibinfo{person}{Mart{\'\i}n Abadi}, \bibinfo{person}{Ashish
  Agarwal}, \bibinfo{person}{Paul Barham}, \bibinfo{person}{Eugene Brevdo},
  \bibinfo{person}{Zhifeng Chen}, \bibinfo{person}{Craig Citro},
  \bibinfo{person}{Greg~S Corrado}, \bibinfo{person}{Andy Davis},
  \bibinfo{person}{Jeffrey Dean}, \bibinfo{person}{Matthieu Devin},
  {et~al\mbox{.}}} \bibinfo{year}{2016}\natexlab{}.
\newblock \showarticletitle{Tensorflow: Large-scale machine learning on
  heterogeneous distributed systems}.
\newblock \bibinfo{journal}{{\em arXiv preprint arXiv:1603.04467\/}}
  (\bibinfo{year}{2016}).
\newblock


\bibitem[\protect\citeauthoryear{Bengio, Courville, and Vincent}{Bengio
  et~al\mbox{.}}{2013}]%
        {bengio2013representation}
\bibfield{author}{\bibinfo{person}{Yoshua Bengio}, \bibinfo{person}{Aaron
  Courville}, {and} \bibinfo{person}{Pascal Vincent}.}
  \bibinfo{year}{2013}\natexlab{}.
\newblock \showarticletitle{Representation learning: A review and new
  perspectives}.
\newblock \bibinfo{journal}{{\em IEEE transactions on pattern analysis and
  machine intelligence\/}} \bibinfo{volume}{35}, \bibinfo{number}{8}
  (\bibinfo{year}{2013}), \bibinfo{pages}{1798--1828}.
\newblock


\bibitem[\protect\citeauthoryear{Bengio and Delalleau}{Bengio and
  Delalleau}{2011}]%
        {bengio2011expressive}
\bibfield{author}{\bibinfo{person}{Yoshua Bengio} {and}
  \bibinfo{person}{Olivier Delalleau}.} \bibinfo{year}{2011}\natexlab{}.
\newblock \showarticletitle{On the expressive power of deep architectures}. In
  \bibinfo{booktitle}{{\em Algorithmic Learning Theory}}. Springer,
  \bibinfo{pages}{18--36}.
\newblock


\bibitem[\protect\citeauthoryear{Berger, Pietra, and Pietra}{Berger
  et~al\mbox{.}}{1996}]%
        {Berger1996A}
\bibfield{author}{\bibinfo{person}{Adam~L. Berger}, \bibinfo{person}{Stephen
  A.~Della Pietra}, {and} \bibinfo{person}{Vincent J.~Della Pietra}.}
  \bibinfo{year}{1996}\natexlab{}.
\newblock \showarticletitle{A Maximum Entropy approach to Natural Language
  Processing}. In \bibinfo{booktitle}{{\em COMPUTATIONAL LINGUISTICS}}.
  \bibinfo{pages}{39--71}.
\newblock


\bibitem[\protect\citeauthoryear{Borisyuk, Zhang, and Kenthapadi}{Borisyuk
  et~al\mbox{.}}{2017}]%
        {borisyuk2017lijar}
\bibfield{author}{\bibinfo{person}{Fedor Borisyuk}, \bibinfo{person}{Liang
  Zhang}, {and} \bibinfo{person}{Krishnaram Kenthapadi}.}
  \bibinfo{year}{2017}\natexlab{}.
\newblock \showarticletitle{LiJAR: A system for job application redistribution
  towards efficient career marketplace}. In \bibinfo{booktitle}{{\em
  Proceedings of the 23rd ACM SIGKDD International Conference on Knowledge
  Discovery and Data Mining}}. ACM, \bibinfo{pages}{1397--1406}.
\newblock


\bibitem[\protect\citeauthoryear{Chen, Shi, Qiu, and Huang}{Chen
  et~al\mbox{.}}{2017}]%
        {chen2017adversarial}
\bibfield{author}{\bibinfo{person}{Xinchi Chen}, \bibinfo{person}{Zhan Shi},
  \bibinfo{person}{Xipeng Qiu}, {and} \bibinfo{person}{Xuanjing Huang}.}
  \bibinfo{year}{2017}\natexlab{}.
\newblock \showarticletitle{Adversarial Multi-Criteria Learning for Chinese
  Word Segmentation}.
\newblock \bibinfo{journal}{{\em NIPS\/}}.
\newblock


\bibitem[\protect\citeauthoryear{Cheng, Koc, Harmsen, Shaked, Chandra, Aradhye,
  Anderson, Corrado, Chai, Ispir, et~al\mbox{.}}{Cheng et~al\mbox{.}}{2016}]%
        {cheng2016wide}
\bibfield{author}{\bibinfo{person}{Heng-Tze Cheng}, \bibinfo{person}{Levent
  Koc}, \bibinfo{person}{Jeremiah Harmsen}, \bibinfo{person}{Tal Shaked},
  \bibinfo{person}{Tushar Chandra}, \bibinfo{person}{Hrishi Aradhye},
  \bibinfo{person}{Glen Anderson}, \bibinfo{person}{Greg Corrado},
  \bibinfo{person}{Wei Chai}, \bibinfo{person}{Mustafa Ispir}, {et~al\mbox{.}}}
  \bibinfo{year}{2016}\natexlab{}.
\newblock \showarticletitle{Wide \& deep learning for recommender systems}. In
  \bibinfo{booktitle}{{\em Proceedings of the 1st Workshop on Deep Learning for
  Recommender Systems}}. ACM, \bibinfo{pages}{7--10}.
\newblock


\bibitem[\protect\citeauthoryear{Collobert, Weston, Bottou, Karlen,
  Kavukcuoglu, and Kuksa}{Collobert et~al\mbox{.}}{2011}]%
        {collobert2011natural}
\bibfield{author}{\bibinfo{person}{Ronan Collobert}, \bibinfo{person}{Jason
  Weston}, \bibinfo{person}{L{\'e}on Bottou}, \bibinfo{person}{Michael Karlen},
  \bibinfo{person}{Koray Kavukcuoglu}, {and} \bibinfo{person}{Pavel Kuksa}.}
  \bibinfo{year}{2011}\natexlab{}.
\newblock \showarticletitle{Natural language processing (almost) from scratch}.
\newblock \bibinfo{journal}{{\em Journal of Machine Learning Research\/}}
  \bibinfo{volume}{12}, \bibinfo{number}{Aug} (\bibinfo{year}{2011}),
  \bibinfo{pages}{2493--2537}.
\newblock


\bibitem[\protect\citeauthoryear{Covington, Adams, and Sargin}{Covington
  et~al\mbox{.}}{2016}]%
        {covington2016deep}
\bibfield{author}{\bibinfo{person}{Paul Covington}, \bibinfo{person}{Jay
  Adams}, {and} \bibinfo{person}{Emre Sargin}.}
  \bibinfo{year}{2016}\natexlab{}.
\newblock \showarticletitle{Deep neural networks for youtube recommendations}.
  In \bibinfo{booktitle}{{\em Proceedings of the 10th ACM Conference on
  Recommender Systems}}. ACM, \bibinfo{pages}{191--198}.
\newblock


\bibitem[\protect\citeauthoryear{Duchi, Hazan, and Singer}{Duchi
  et~al\mbox{.}}{2011}]%
        {duchi2011adaptive}
\bibfield{author}{\bibinfo{person}{John Duchi}, \bibinfo{person}{Elad Hazan},
  {and} \bibinfo{person}{Yoram Singer}.} \bibinfo{year}{2011}\natexlab{}.
\newblock \showarticletitle{Adaptive subgradient methods for online learning
  and stochastic optimization}.
\newblock \bibinfo{journal}{{\em Journal of Machine Learning Research\/}}
  \bibinfo{volume}{12}, \bibinfo{number}{Jul} (\bibinfo{year}{2011}),
  \bibinfo{pages}{2121--2159}.
\newblock


\bibitem[\protect\citeauthoryear{Gopinath and Strickman}{Gopinath and
  Strickman}{2010}]%
        {gopinath2010personalized}
\bibfield{author}{\bibinfo{person}{Dinesh Gopinath} {and}
  \bibinfo{person}{Michael Strickman}.} \bibinfo{year}{2010}\natexlab{}.
\newblock \bibinfo{title}{Personalized advertising and recommendation}.
\newblock   (\bibinfo{date}{Aug.~30} \bibinfo{year}{2010}).
\newblock
\newblock
\shownote{US Patent App. 12/871,416.}


\bibitem[\protect\citeauthoryear{Hidasi, Karatzoglou, Baltrunas, and
  Tikk}{Hidasi et~al\mbox{.}}{2015}]%
        {hidasi2015session}
\bibfield{author}{\bibinfo{person}{Bal{\'a}zs Hidasi},
  \bibinfo{person}{Alexandros Karatzoglou}, \bibinfo{person}{Linas Baltrunas},
  {and} \bibinfo{person}{Domonkos Tikk}.} \bibinfo{year}{2015}\natexlab{}.
\newblock \showarticletitle{Session-based recommendations with recurrent neural
  networks}.
\newblock \bibinfo{journal}{{\em arXiv preprint arXiv:1511.06939\/}}
  (\bibinfo{year}{2015}).
\newblock


\bibitem[\protect\citeauthoryear{Hochreiter and Schmidhuber}{Hochreiter and
  Schmidhuber}{1997}]%
        {Hochreiter}
\bibfield{author}{\bibinfo{person}{Sepp Hochreiter} {and}
  \bibinfo{person}{J\"{u}rgen Schmidhuber}.} \bibinfo{year}{1997}\natexlab{}.
\newblock \showarticletitle{Long Short-Term Memory}.
\newblock \bibinfo{journal}{{\em Neural Comput.\/}} \bibinfo{volume}{9},
  \bibinfo{number}{8} (\bibinfo{date}{Nov.} \bibinfo{year}{1997}),
  \bibinfo{pages}{1735--1780}.
\newblock
\showISSN{0899-7667}


\bibitem[\protect\citeauthoryear{Huang, He, Gao, Deng, Acero, and Heck}{Huang
  et~al\mbox{.}}{2013}]%
        {huang2013}
\bibfield{author}{\bibinfo{person}{Po-Sen Huang}, \bibinfo{person}{Xiaodong
  He}, \bibinfo{person}{Jianfeng Gao}, \bibinfo{person}{Li Deng},
  \bibinfo{person}{Alex Acero}, {and} \bibinfo{person}{Larry Heck}.}
  \bibinfo{year}{2013}\natexlab{}.
\newblock \showarticletitle{Learning deep structured semantic models for web
  search using clickthrough data}. In \bibinfo{booktitle}{{\em Proceedings of
  the 22nd ACM international conference on Conference on information \&
  knowledge management}}. ACM, \bibinfo{pages}{2333--2338}.
\newblock


\bibitem[\protect\citeauthoryear{Koren, Bell, and Volinsky}{Koren
  et~al\mbox{.}}{2009}]%
        {koren2009matrix}
\bibfield{author}{\bibinfo{person}{Yehuda Koren}, \bibinfo{person}{Robert
  Bell}, {and} \bibinfo{person}{Chris Volinsky}.}
  \bibinfo{year}{2009}\natexlab{}.
\newblock \showarticletitle{Matrix factorization techniques for recommender
  systems}.
\newblock \bibinfo{journal}{{\em Computer\/}} \bibinfo{volume}{42},
  \bibinfo{number}{8} (\bibinfo{year}{2009}).
\newblock


\bibitem[\protect\citeauthoryear{Krizhevsky, Sutskever, and Hinton}{Krizhevsky
  et~al\mbox{.}}{2012}]%
        {krizhevsky2012imagenet}
\bibfield{author}{\bibinfo{person}{Alex Krizhevsky}, \bibinfo{person}{Ilya
  Sutskever}, {and} \bibinfo{person}{Geoffrey~E Hinton}.}
  \bibinfo{year}{2012}\natexlab{}.
\newblock \showarticletitle{Imagenet classification with deep convolutional
  neural networks}. In \bibinfo{booktitle}{{\em Advances in neural information
  processing systems}}. \bibinfo{pages}{1097--1105}.
\newblock


\bibitem[\protect\citeauthoryear{Linden, Smith, and York}{Linden
  et~al\mbox{.}}{2003}]%
        {linden2003amazon}
\bibfield{author}{\bibinfo{person}{Greg Linden}, \bibinfo{person}{Brent Smith},
  {and} \bibinfo{person}{Jeremy York}.} \bibinfo{year}{2003}\natexlab{}.
\newblock \showarticletitle{Amazon. com recommendations: Item-to-item
  collaborative filtering}.
\newblock \bibinfo{journal}{{\em IEEE Internet computing\/}}
  \bibinfo{volume}{7}, \bibinfo{number}{1} (\bibinfo{year}{2003}),
  \bibinfo{pages}{76--80}.
\newblock


\bibitem[\protect\citeauthoryear{Liu, Gao, He, Deng, Duh, and Wang}{Liu
  et~al\mbox{.}}{2015}]%
        {liu2015representation}
\bibfield{author}{\bibinfo{person}{Xiaodong Liu}, \bibinfo{person}{Jianfeng
  Gao}, \bibinfo{person}{Xiaodong He}, \bibinfo{person}{Li Deng},
  \bibinfo{person}{Kevin Duh}, {and} \bibinfo{person}{Ye-Yi Wang}.}
  \bibinfo{year}{2015}\natexlab{}.
\newblock \showarticletitle{Representation Learning Using Multi-Task Deep
  Neural Networks for Semantic Classification and Information Retrieval.}. In
  \bibinfo{booktitle}{{\em HLT-NAACL}}. \bibinfo{pages}{912--921}.
\newblock


\bibitem[\protect\citeauthoryear{McMahan}{McMahan}{2011}]%
        {mcmahan2011follow}
\bibfield{author}{\bibinfo{person}{Brendan McMahan}.}
  \bibinfo{year}{2011}\natexlab{}.
\newblock \showarticletitle{Follow-the-regularized-leader and mirror descent:
  Equivalence theorems and l1 regularization}. In \bibinfo{booktitle}{{\em
  Proceedings of the Fourteenth International Conference on Artificial
  Intelligence and Statistics}}. \bibinfo{pages}{525--533}.
\newblock


\bibitem[\protect\citeauthoryear{Mikolov, Corrado, Chen, Dean, Mikolov,
  Corrado, Chen, and Dean}{Mikolov et~al\mbox{.}}{2013}]%
        {Mikolov2013Efficient}
\bibfield{author}{\bibinfo{person}{Tomas Mikolov}, \bibinfo{person}{Greg
  Corrado}, \bibinfo{person}{Kai Chen}, \bibinfo{person}{Jeffrey Dean},
  \bibinfo{person}{Tomas Mikolov}, \bibinfo{person}{Greg Corrado},
  \bibinfo{person}{Kai Chen}, {and} \bibinfo{person}{Jeffrey Dean}.}
  \bibinfo{year}{2013}\natexlab{}.
\newblock \showarticletitle{Efficient Estimation of Word Representations in
  Vector Space}. In \bibinfo{booktitle}{{\em International Conference on
  Learning Representations}}. \bibinfo{pages}{1--12}.
\newblock


\bibitem[\protect\citeauthoryear{Oh, Lee, Lim, and Choi}{Oh
  et~al\mbox{.}}{2014}]%
        {oh2014personalized}
\bibfield{author}{\bibinfo{person}{Kyo-Joong Oh}, \bibinfo{person}{Won-Jo Lee},
  \bibinfo{person}{Chae-Gyun Lim}, {and} \bibinfo{person}{Ho-Jin Choi}.}
  \bibinfo{year}{2014}\natexlab{}.
\newblock \showarticletitle{Personalized news recommendation using classified
  keywords to capture user preference}. In \bibinfo{booktitle}{{\em Advanced
  Communication Technology (ICACT), 2014 16th International Conference on}}.
  IEEE, \bibinfo{pages}{1283--1287}.
\newblock


\bibitem[\protect\citeauthoryear{Ramsundar, Kearnes, Riley, Webster, Konerding,
  and Pande}{Ramsundar et~al\mbox{.}}{2015}]%
        {ramsundar2015massively}
\bibfield{author}{\bibinfo{person}{Bharath Ramsundar}, \bibinfo{person}{Steven
  Kearnes}, \bibinfo{person}{Patrick Riley}, \bibinfo{person}{Dale Webster},
  \bibinfo{person}{David Konerding}, {and} \bibinfo{person}{Vijay Pande}.}
  \bibinfo{year}{2015}\natexlab{}.
\newblock \showarticletitle{Massively multitask networks for drug discovery}.
\newblock \bibinfo{journal}{{\em arXiv preprint arXiv:1502.02072\/}}
  (\bibinfo{year}{2015}).
\newblock


\bibitem[\protect\citeauthoryear{Ranjan, Patel, and Chellappa}{Ranjan
  et~al\mbox{.}}{2016}]%
        {ranjan2016hyperface}
\bibfield{author}{\bibinfo{person}{Rajeev Ranjan}, \bibinfo{person}{Vishal~M
  Patel}, {and} \bibinfo{person}{Rama Chellappa}.}
  \bibinfo{year}{2016}\natexlab{}.
\newblock \showarticletitle{Hyperface: A deep multi-task learning framework for
  face detection, landmark localization, pose estimation, and gender
  recognition}.
\newblock \bibinfo{journal}{{\em arXiv preprint arXiv:1603.01249\/}}
  (\bibinfo{year}{2016}).
\newblock


\bibitem[\protect\citeauthoryear{Salakhutdinov, Mnih, and Hinton}{Salakhutdinov
  et~al\mbox{.}}{2007}]%
        {Salakhutdinov2007}
\bibfield{author}{\bibinfo{person}{Ruslan Salakhutdinov},
  \bibinfo{person}{Andriy Mnih}, {and} \bibinfo{person}{Geoffrey Hinton}.}
  \bibinfo{year}{2007}\natexlab{}.
\newblock \showarticletitle{Restricted Boltzmann machines for collaborative
  filtering}. In \bibinfo{booktitle}{{\em Proceedings of the 24th international
  conference on Machine learning}}. ACM, \bibinfo{pages}{791--798}.
\newblock


\bibitem[\protect\citeauthoryear{Sedhain, Menon, Sanner, and Xie}{Sedhain
  et~al\mbox{.}}{2015}]%
        {Suvash2016}
\bibfield{author}{\bibinfo{person}{Suvash Sedhain},
  \bibinfo{person}{Aditya~Krishna Menon}, \bibinfo{person}{Scott Sanner}, {and}
  \bibinfo{person}{Lexing Xie}.} \bibinfo{year}{2015}\natexlab{}.
\newblock \showarticletitle{Autorec: Autoencoders meet collaborative
  filtering}. In \bibinfo{booktitle}{{\em Proceedings of the 24th International
  Conference on World Wide Web}}. ACM, \bibinfo{pages}{111--112}.
\newblock


\bibitem[\protect\citeauthoryear{Seltzer and Droppo}{Seltzer and
  Droppo}{2013}]%
        {seltzer2013multi}
\bibfield{author}{\bibinfo{person}{Michael~L Seltzer} {and}
  \bibinfo{person}{Jasha Droppo}.} \bibinfo{year}{2013}\natexlab{}.
\newblock \showarticletitle{Multi-task learning in deep neural networks for
  improved phoneme recognition}. In \bibinfo{booktitle}{{\em Acoustics, Speech
  and Signal Processing (ICASSP), 2013 IEEE International Conference on}}.
  IEEE, \bibinfo{pages}{6965--6969}.
\newblock


\bibitem[\protect\citeauthoryear{Srivastava, Hinton, Krizhevsky, Sutskever, and
  Salakhutdinov}{Srivastava et~al\mbox{.}}{2014}]%
        {srivastava2014dropout}
\bibfield{author}{\bibinfo{person}{Nitish Srivastava},
  \bibinfo{person}{Geoffrey~E Hinton}, \bibinfo{person}{Alex Krizhevsky},
  \bibinfo{person}{Ilya Sutskever}, {and} \bibinfo{person}{Ruslan
  Salakhutdinov}.} \bibinfo{year}{2014}\natexlab{}.
\newblock \showarticletitle{Dropout: a simple way to prevent neural networks
  from overfitting.}
\newblock \bibinfo{journal}{{\em Journal of machine learning research\/}}
  \bibinfo{volume}{15}, \bibinfo{number}{1} (\bibinfo{year}{2014}),
  \bibinfo{pages}{1929--1958}.
\newblock


\bibitem[\protect\citeauthoryear{Tan, Xu, and Liu}{Tan et~al\mbox{.}}{2016}]%
        {tan2016improved}
\bibfield{author}{\bibinfo{person}{Yong~Kiam Tan}, \bibinfo{person}{Xinxing
  Xu}, {and} \bibinfo{person}{Yong Liu}.} \bibinfo{year}{2016}\natexlab{}.
\newblock \showarticletitle{Improved recurrent neural networks for
  session-based recommendations}. In \bibinfo{booktitle}{{\em Proceedings of
  the 1st Workshop on Deep Learning for Recommender Systems}}. ACM,
  \bibinfo{pages}{17--22}.
\newblock


\bibitem[\protect\citeauthoryear{Ustinovskiy, Gusev, and Serdyukov}{Ustinovskiy
  et~al\mbox{.}}{2015}]%
        {ustinovskiy2015optimization}
\bibfield{author}{\bibinfo{person}{Yury Ustinovskiy}, \bibinfo{person}{Gleb
  Gusev}, {and} \bibinfo{person}{Pavel Serdyukov}.}
  \bibinfo{year}{2015}\natexlab{}.
\newblock \showarticletitle{An optimization framework for weighting implicit
  relevance labels for personalized web search}. In \bibinfo{booktitle}{{\em
  Proceedings of the 24th International Conference on World Wide Web}}.
  International World Wide Web Conferences Steering Committee,
  \bibinfo{pages}{1144--1154}.
\newblock


\bibitem[\protect\citeauthoryear{Van~den Oord, Dieleman, and Schrauwen}{Van~den
  Oord et~al\mbox{.}}{2013}]%
        {Oord2013}
\bibfield{author}{\bibinfo{person}{Aaron Van~den Oord}, \bibinfo{person}{Sander
  Dieleman}, {and} \bibinfo{person}{Benjamin Schrauwen}.}
  \bibinfo{year}{2013}\natexlab{}.
\newblock \showarticletitle{Deep content-based music recommendation}. In
  \bibinfo{booktitle}{{\em Advances in neural information processing systems}}.
  \bibinfo{pages}{2643--2651}.
\newblock


\bibitem[\protect\citeauthoryear{Wang, Wang, and Yeung}{Wang
  et~al\mbox{.}}{2015}]%
        {wanghao2015}
\bibfield{author}{\bibinfo{person}{Hao Wang}, \bibinfo{person}{Naiyan Wang},
  {and} \bibinfo{person}{Dit-Yan Yeung}.} \bibinfo{year}{2015}\natexlab{}.
\newblock \showarticletitle{Collaborative deep learning for recommender
  systems}. In \bibinfo{booktitle}{{\em Proceedings of the 21th ACM SIGKDD
  International Conference on Knowledge Discovery and Data Mining}}. ACM,
  \bibinfo{pages}{1235--1244}.
\newblock


\bibitem[\protect\citeauthoryear{Wang, Bendersky, Metzler, and Najork}{Wang
  et~al\mbox{.}}{2016}]%
        {wang2016learning}
\bibfield{author}{\bibinfo{person}{Xuanhui Wang}, \bibinfo{person}{Michael
  Bendersky}, \bibinfo{person}{Donald Metzler}, {and} \bibinfo{person}{Marc
  Najork}.} \bibinfo{year}{2016}\natexlab{}.
\newblock \showarticletitle{Learning to rank with selection bias in personal
  search}. In \bibinfo{booktitle}{{\em Proceedings of the 40th international
  ACM SIGIR conference on Research and development in Information Retrieval}}.
  ACM, \bibinfo{pages}{115--124}.
\newblock


\bibitem[\protect\citeauthoryear{Wu, DuBois, Zheng, and Ester}{Wu
  et~al\mbox{.}}{2016}]%
        {yaowu2016}
\bibfield{author}{\bibinfo{person}{Yao Wu}, \bibinfo{person}{Christopher
  DuBois}, \bibinfo{person}{Alice~X Zheng}, {and} \bibinfo{person}{Martin
  Ester}.} \bibinfo{year}{2016}\natexlab{}.
\newblock \showarticletitle{Collaborative denoising auto-encoders for top-n
  recommender systems}. In \bibinfo{booktitle}{{\em Proceedings of the Ninth
  ACM International Conference on Web Search and Data Mining}}. ACM,
  \bibinfo{pages}{153--162}.
\newblock


\bibitem[\protect\citeauthoryear{Zhai, Chang, Zhang, and Zhang}{Zhai
  et~al\mbox{.}}{2016}]%
        {zhai2016deepintent}
\bibfield{author}{\bibinfo{person}{Shuangfei Zhai}, \bibinfo{person}{Keng-hao
  Chang}, \bibinfo{person}{Ruofei Zhang}, {and} \bibinfo{person}{Zhongfei~Mark
  Zhang}.} \bibinfo{year}{2016}\natexlab{}.
\newblock \showarticletitle{Deepintent: Learning attentions for online
  advertising with recurrent neural networks}. In \bibinfo{booktitle}{{\em
  Proceedings of the 22nd ACM SIGKDD International Conference on Knowledge
  Discovery and Data Mining}}. ACM, \bibinfo{pages}{1295--1304}.
\newblock


\bibitem[\protect\citeauthoryear{Zhang, Zhang, Zhang, Lai, Liu, Zhang, and
  Ma}{Zhang et~al\mbox{.}}{2015}]%
        {zhang2015daily}
\bibfield{author}{\bibinfo{person}{Yongfeng Zhang}, \bibinfo{person}{Min
  Zhang}, \bibinfo{person}{Yi Zhang}, \bibinfo{person}{Guokun Lai},
  \bibinfo{person}{Yiqun Liu}, \bibinfo{person}{Honghui Zhang}, {and}
  \bibinfo{person}{Shaoping Ma}.} \bibinfo{year}{2015}\natexlab{}.
\newblock \showarticletitle{Daily-aware personalized recommendation based on
  feature-level time series analysis}. In \bibinfo{booktitle}{{\em Proceedings
  of the 24th international conference on world wide web}}. International World
  Wide Web Conferences Steering Committee, \bibinfo{pages}{1373--1383}.
\newblock


\bibitem[\protect\citeauthoryear{Zhang, Luo, Loy, and Tang}{Zhang
  et~al\mbox{.}}{2014}]%
        {zhang2014facial}
\bibfield{author}{\bibinfo{person}{Zhanpeng Zhang}, \bibinfo{person}{Ping Luo},
  \bibinfo{person}{Chen~Change Loy}, {and} \bibinfo{person}{Xiaoou Tang}.}
  \bibinfo{year}{2014}\natexlab{}.
\newblock \showarticletitle{Facial landmark detection by deep multi-task
  learning}. In \bibinfo{booktitle}{{\em European Conference on Computer
  Vision}}. Springer, \bibinfo{pages}{94--108}.
\newblock


\bibitem[\protect\citeauthoryear{Zheng, Noroozi, and Yu}{Zheng
  et~al\mbox{.}}{2017}]%
        {zheng2017joint}
\bibfield{author}{\bibinfo{person}{Lei Zheng}, \bibinfo{person}{Vahid Noroozi},
  {and} \bibinfo{person}{Philip~S Yu}.} \bibinfo{year}{2017}\natexlab{}.
\newblock \showarticletitle{Joint deep modeling of users and items using
  reviews for recommendation}. In \bibinfo{booktitle}{{\em Proceedings of the
  Tenth ACM International Conference on Web Search and Data Mining}}. ACM,
  \bibinfo{pages}{425--434}.
\newblock


\end{thebibliography}

\end{document}